\documentclass[a4paper]{article}

\usepackage[nonatbib,final]{neurips_2022}

\usepackage[utf8]{inputenc} 
\usepackage[T1]{fontenc}    
\usepackage{hyperref}       
\usepackage{url}            
\usepackage{booktabs}       
\usepackage{amsfonts}       
\usepackage{nicefrac}       
\usepackage{microtype}      
\usepackage[dvipsnames]{xcolor}         
\usepackage{amsmath}
\usepackage{graphicx}
\usepackage{tcolorbox}
\usepackage{enumitem}

\newcommand{\Null} {\operatorname{Null}}
\newcommand{\W} {\overline{W}}
\newcommand{\Qspace} {\mathcal{Q}}

\title{The Phenomenon of Policy Churn}

\author{%
  Tom Schaul\\
  DeepMind\\
  London, UK 
 \vspace{-1em}
  \And
  Andr\'{e} Barreto\\
  DeepMind\\
  London, UK 
 \vspace{-1em}
  \And
  John Quan\\
  DeepMind\\
  London, UK 
 \vspace{-1em}
  \And
  Georg Ostrovski\\
  DeepMind\\
  London, UK 
 \vspace{-1em}
  \AND
  \centerline{
  \texttt{\{tom,andrebarreto,johnquan,ostrovski\}@deepmind.com}}\
}

\begin{document}

\maketitle

\begin{abstract}
We identify and study the phenomenon of \emph{policy churn}, that is, the rapid change of the greedy policy in value-based reinforcement learning. Policy churn operates at a \emph{surprisingly rapid} pace, changing the greedy action in a large fraction of states within a handful of learning updates (in a typical deep RL set-up such as DQN on Atari).
We characterise the phenomenon empirically, verifying that it is not limited to specific algorithm or environment properties. 
A number of ablations help whittle down the plausible explanations on why churn occurs, the most likely one being deep learning with high-variance updates. Finally, we hypothesise that policy churn is a potentially beneficial but overlooked form of \emph{implicit exploration}, which casts $\epsilon$-greedy exploration in a fresh light, namely that $\epsilon$-noise plays a much smaller role than expected.
 \vspace{-1em}
\end{abstract}

\section{The Phenomenon}
Reinforcement learning (RL) involves agents that incrementally update their policy. 
This process is driven by the objective of maximising reward, and based on experience that the agent generates via exploration.
The sequence of policies $\pi_0, \ldots, \pi_k, \ldots, \pi_T$ usually starts from a randomly initialised policy $\pi_0$ and aims to end at a near-optimal policy $\pi_T \approx \pi^{*}$.
Ideally, steps in that sequence ($\pi_k \rightarrow \pi_{k+1}$) are policy improvements that increase expected reward. 

This paper studies the amount of \emph{policy change} that goes along with such a policy update process (for a definition, see Section~\ref{sec:def}).
In particular, it makes the core observation that policy change \emph{in practice} (as illustrated in some typical deep RL settings)
is orders of magnitude larger than could have been expected, and stands in contrast to  various reference algorithms (Sections~\ref{sec:quantifying} and \ref{sec:catch}).

\begin{figure}[tb]
    \centering
    \includegraphics[width=0.8\textwidth]{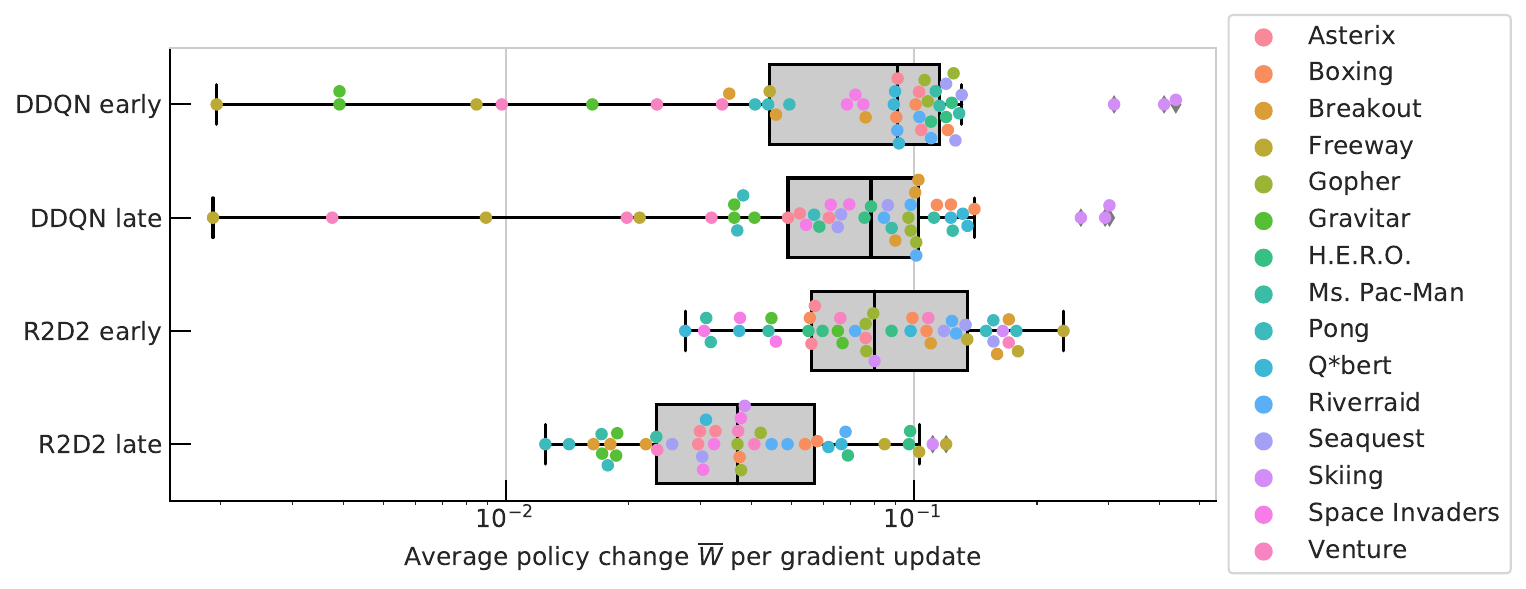}
    \caption{Average amount of policy change $\W$ (Eq.~\ref{eq:w-bar}) per update, in two deep RL agents (DoubleDQN and R2D2). 
    Points are averages over $3$ seeds, on one of $15$ (colour-coded) Atari games. ``Early'' denotes the first $25\%$ of training, ``late'' denotes the final $25\%$; observe that churn magnitudes drop late in training, but only for R2D2.
    See Figures~\ref{fig:typical-churn} and~\ref{fig:r2d2-churn} in the appendix for more fine-grained results.
}
    \label{fig:summary_churn}
\end{figure}

\begin{tcolorbox}[leftrule=1.5mm,top=1mm,bottom=0mm]
\textbf{Key observation 1:}
The greedy policy changes much more rapidly than you probably think.\footnote{As a coarse magnitude for the impatient reader: in a typical run of DQN on Atari, the greedy policy changes in $\approx10\%$ of all states after \emph{a single gradient update} (Figure~\ref{fig:summary_churn} and Section~\ref{sec:quantifying}).}
\end{tcolorbox}

We dub this phenomenon ``\emph{policy churn}'' to highlight that most of this policy change may be unnecessary.
We study the phenomenon in depth, determining the range of deep RL scenarios it appears in, fleshing out its properties, and in the process narrowing the space of potential causes and mechanisms involved using a set of ablations (Section~\ref{sec:causes}).

Our second key message relates the phenomenon of churn to exploration, 
specifically in the context of $\epsilon$-greedy exploration (Section~\ref{sec:exploration}), 
with some more speculative ramifications in Section~\ref{sec:consequences}.

\begin{tcolorbox}[leftrule=1.5mm,top=1mm,bottom=0mm]
\textbf{Key observation 2:}
Policy churn is a significant driver of exploration.\footnote{
This holds both in the sense that reducing churn can reduce performance, and in the sense that explicitly adding noise becomes unnecessary in the presence of churn (i.e., $\epsilon=0$ is viable).}
\end{tcolorbox}

\subsection{Defining policy change}
\label{sec:def}
A policy is a function from states $s \in \mathcal{S}$ to a distribution over actions $a \in \mathcal{A}$, where for the purposes of this paper $\mathcal{A}$ is discrete.
We quantify the local, per-state \emph{policy change} between policies $\pi$ and $\pi'$ using the moved probability mass (i.e., the total variation distance): 
\begin{equation}
\label{eq:pi-change}
W(\pi, \pi' |s) : = \frac{1}{2}\sum_a \left|\pi(a|s) - \pi'(a|s)\right|,
\end{equation}
which satisfies $0 \leq W(\pi, \pi' |s) \leq 1$.
When $\pi$ and $\pi'$ are \emph{greedy} policies derived from a state-action-value function---that is, $\pi(s)\in\arg\max_a q(s,a)$---then 
$W(\pi, \pi' |s) = 1$ if the $\arg\max$ action in state $s$ changes upon replacing $q$
by the function $q'$ underlying $\pi'$, and $W(\pi, \pi' |s) = 0$ otherwise.
Similar reasoning applies more generally when both $\pi$ and $\pi'$ are deterministic policies.
We can aggregate policy change across all states, weighted by a state-distribution $d$:
\begin{equation}
    \W(\pi, \pi') := \mathbb{E}_{s\sim d}[W(\pi, \pi' |s)],
\label{eq:w-bar}
\end{equation}
where a reasonable choice for $d$ is the empirical state distribution encountered during training (which is non-stationary, but it could also be the stationary distribution of a fixed policy, or the uniform distribution across all states, as discussed for some of the toy scenarios below).
For greedy policies (and uniform $d$), $\W(\pi, \pi')$ is simply the fraction of states where an $\arg\max$ switch occurred.

For settings where policy performance stabilises at some point $t=P$ of training (e.g., hitting a performance plateau, or converging to optimal behaviour), two additional metrics may be of interest, namely the \emph{cumulative} policy change $\W_{0:P}$ until that point\footnote{
Note that the granularity of updates (e.g., batch size) determines how many intermediate policies are considered, which in turn affects the measured magnitude of policy change in learning processes that have churn.
},
and the \emph{average} policy change after that point $\W^{+}$ (which could be zero, e.g., if the process converges):
\begin{eqnarray}
\W_{0:P} := \sum_{i=0}^{P-1}\W(\pi_{i}, \pi_{i+1})
&\quad\quad\quad&
\W^{+} := \limsup_{T\rightarrow\infty} \frac{1}{T-P}\W_{P:T}
\,.
\label{eq:conv-churn}
\end{eqnarray}

\subsection{Quantifying the phenomenon}
\label{sec:quantifying}

Given an initial and a final policy, the process with the minimum amount of policy change is an oracle that jumps from $\pi_0$ to $\pi_P$ in a single step ($P=1$). By construction, this can incur a policy change of at most one unit, $\W_{0:P} \leq 1$. 
In value-based deep RL agents, a natural definition for the sequence of policies is to use the induced greedy policies $\pi_k(s) \in \arg\max_a q_{\theta_{k}}(s,a)$
where $\theta_{k}$ are the parameters of the Q-function at iteration $k$. In agents that use a target network (inducing $\pi$) that is an older copy of the online network (inducing $\pi'$), it is easy to measure $\W(\pi, \pi')$ by comparing their $\arg\max$ actions at the points in training where the target network lags behind by just one update.
Figure~\ref{fig:summary_churn} shows typical values for $\W$ on a few Atari games, estimated by comparing the policies induced by online and (one update old) target networks, on batches of experience sampled from the agent's replay buffer. It is worth emphasising that with such rates of average policy change, of $\W\approx10\%$ per update, the magnitude of whole-lifetime change becomes enormous: across training, an agent like DQN, which performs $10^7$ updates, changes its greedy action \emph{a million times in each state} (on average).

A second striking result is the amount of policy change in \emph{late} training, when the performance of the policy no longer changes (in the case of \textsc{Pong} it is arguably optimal): there is still a change of $\W^+\approx5\%$ per update (Figures~\ref{fig:summary_churn} and~\ref{fig:pong-late}). This highlights that a lot (maybe most) of policy churn is not directed at a policy improvement; we revisit this in Section~\ref{sec:null}.

\paragraph{Is this unexpected?}
\label{sec:survey}
We conducted an informal survey of over 50 deep RL practitioners, including three of the inventors of DQN~\cite{dqn-arxiv}, for their estimate on how rapidly the policy changes in a typical Q-learning based setup. The question was: \texttt{For the greedy policy to change in 10\% of all states, how many learning updates does it take?} (or equivalent).
The median response was $1\,000$ updates, with answers varying between $1$ and $1\,000\,000$ updates. This deviates by three orders of magnitude from the empirical value of $1$ update (or $\W\approx 0.1$, see Figure~\ref{fig:summary_churn}).

\paragraph{Policy change in other settings.}
To appreciate the large empirical magnitudes of (cumulative) policy change we observe in deep RL, we can contrast them with a few alternative settings.
For example, classic dynamic programming techniques such as value iteration or policy iteration~\cite{suttonbook}, when applied to toy RL domains such as FourRooms, \textsc{Catch}, or DeepSea~\cite{bsuite}, accumulate $\W_{0:P} \approx 1$, not much more than the single-step oracle (see Appendix~\ref{app:toy-mdp}). The two main differences from deep RL are their tabular nature (no function approximation, FA) and non-incremental updates.
It is possible to construct tabular settings with much larger policy change, with either incremental updates (Appendix~\ref{app:2-arm-bandit}) or bootstrapping (Appendix~\ref{app:adversarial-td}).
A minimalist example with non-linear FA and incremental updates is supervised learning on the MNIST dataset (the ``policy'' in this case are the predicted label probabilities). Training a digit classifier to convergence accumulates $\W_{0:P} \approx 10$, that is, the average input goes through $10$ label switches (see Appendix~\ref{app:mnist}).
None of these examples are fully satisfying, as they are not apples-to-apples comparisons; so in Section~\ref{sec:catch} we construct a spectrum of algorithmic variants that spans from tabular policy iteration (without churn), via tabular Q-learning to an approximation of DQN (with realistic magnitudes of churn).

\begin{figure}[tb]
    \centering
    \includegraphics[width=\textwidth]{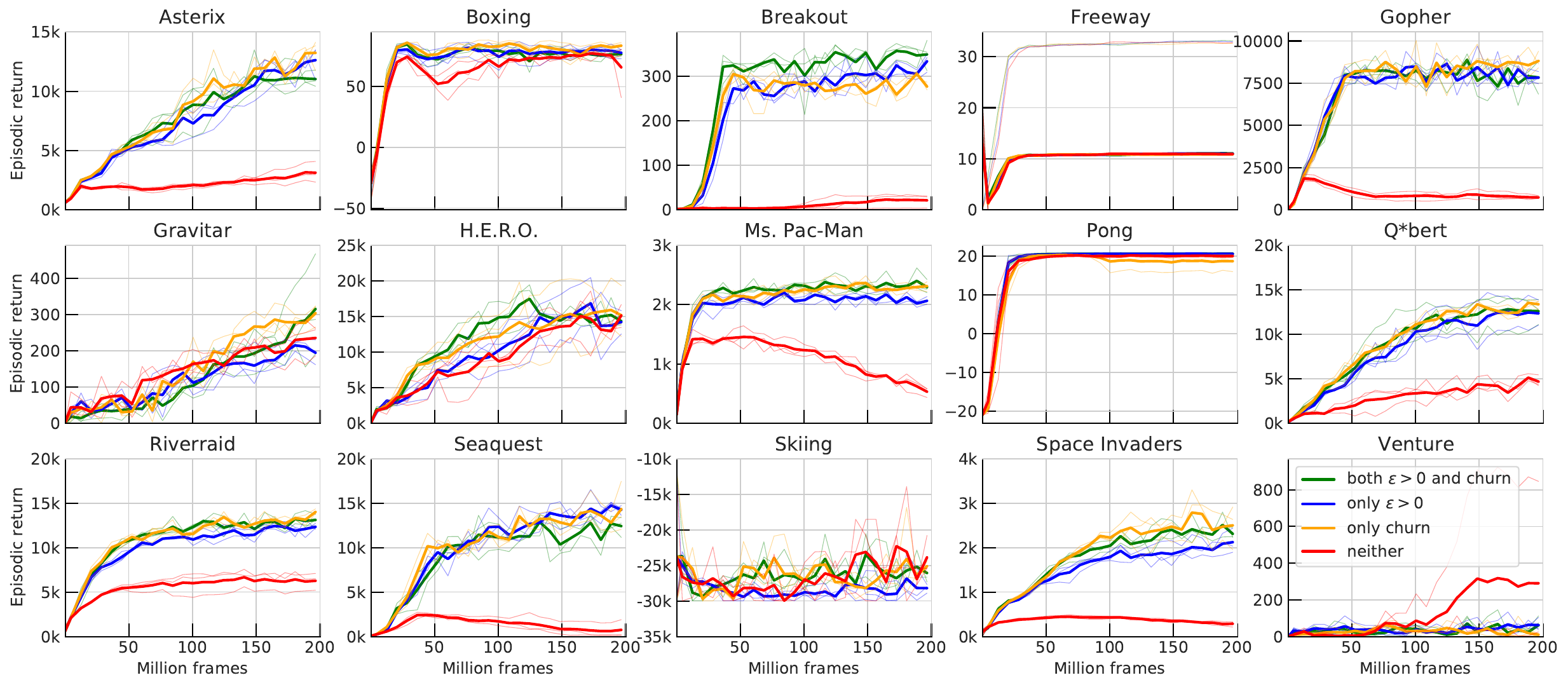}
    \caption{Impact of churn on exploration. Plots show performance of DoubleDQN on Atari, with four variants of exploration:
    {\color{OliveGreen} \bf green} is the unmodified baseline ($\epsilon=0.01$), {\color{Dandelion} \bf gold} changes $\epsilon=0$, {\color{blue}\bf blue} has reduced churn by acting with the target network, and {\color{red}\bf red} shows the effect of both of these changes together (no more exploration, and a corresponding performance collapse in most games).
    Thin lines depict individual seeds (3 per setting), thick lines are averages, smoothed over 10M frames.}
    \label{fig:zero-eps}
\end{figure}

\section{The Exploration Effect}
\label{sec:exploration}
We now turn to the impact of policy churn on agent behaviour:
What happens when \emph{acting} according to policies produced by a learning process that induces rapidly changing policies? In other words, what is the effect of policy churn on exploration?
While each individual greedy policy would lead to a very narrow set of experience in a (nearly) deterministic environment like Atari, the fact that the greedy policy changes so rapidly (in $10\%$ of states per update, with one update every $16$ environment frames in DoubleDQN) makes for a broad data distribution.
And in many circumstances this is sufficient for good performance, even in the absence of any other form of exploration, such as stochasticity introduced via an $\epsilon$-greedy policy. 
Figure~\ref{fig:zero-eps} shows this across a range of Atari games: compare {\color{OliveGreen}\bf green} (baseline) and {\color{Dandelion}\bf gold} ($\epsilon=0$) curves.\footnote{
In line with prior work, all our DoubleDQN experiments preserve the decaying $\epsilon$-schedule in the first $2\%$ of training, it is only zero after that initial phase; R2D2 experiments have no such schedule.}

Conversely, removing (some of the) churn in the behaviour during training, which can be done by acting with the target network (updated only every $120\,000$ frames in the DoubleDQN agent) instead of the online network, sometimes reduces performance even in the presence of $\epsilon$-greedy exploration ({\color{blue}\bf blue} in Figure~\ref{fig:zero-eps}).\footnote{
Acting with the target network also introduces \emph{latency} on how fast newly learned knowledge can be exploited. To see how specifically this latency should have a negligible effect on performance, imagine shifting the $x$-axis of the blue curve by $120\,000$ frames to the left, which would be an oracle ``target-network-of-the-future'' variant.
}
Additionally, we show that performance often \emph{collapses} completely when both forms of exploration are removed ($\epsilon =0$ and no churn, in {\color{red}\bf red}).
Figure~\ref{fig:zero-eps} compares all four variants of exploration, indicating that the two sources of exploration have different contributions in different games.

\paragraph{Sufficient exploration with $\epsilon=0$.}
The perhaps unintuitive observation of successful exploration with purely greedy policies has been made before, albeit implicitly. In particular, in the presence of
certain alternative exploration methods such as noisy nets \cite{fortunato2017noisy}, no significant additional advantage is obtained
from using $\epsilon > 0$ \cite{rainbow}. Other works containing experimental variants with $\epsilon = 0$ demonstrated successful training
in this setting \cite{pislar2021agents,adx-bandit-arxiv}, but did not highlight the result.

\paragraph{Consistent behaviour and Thompson sampling.}
In considering the potential exploration benefits of a rapidly changing policy, it is worth qualitatively contrasting the resulting behaviour with that of an $\epsilon$-greedy policy with $\epsilon > 0$. The latter generates high-frequency dithering, with uncorrelated random action decisions at consecutive states and the effect of most exploratory actions likely undone by the following greedy action \cite{ezgreedy}.
By contrast, policy-churn-induced exploration can be expected to generate temporally correlated, consistent exploration (necessary, though perhaps not sufficient, to perform ``deep'' exploration, as in ensemble and Thompson sampling methods \cite{thompson1933likelihood,bootstrap_dqn}). On the other hand, while $\epsilon$-greedy exploration is explicitly unbiased in action space, policy churn likely prefers exploration across near-optimal actions (with respect to the current value function).
This may be beneficial in some settings, for example when some actions are deadly, and high $\epsilon$ prevents long episodes. It may also be detrimental in others, where $\epsilon > 0$ helps the agent get unstuck.

\section{Potential Causes}
\label{sec:causes}

With the presence and impact of the churn phenomenon established, this section aims to provide additional depth.
First, we look at the generality of the effect in Section~\ref{sec:range}. Second, we conduct investigations into the sensitivity of the phenomenon, with Section~\ref{sec:ablations} showing ablations to the large-scale deep RL agents (more material in Appendix~\ref{app:more-ablations}), and Section~\ref{sec:catch} taking the complementary approach of interpolating between dynamic programming and a DQN approximation on a toy domain.
Finally, Section~\ref{sec:hypothesis} synthesises the findings and postulates some compatible underlying mechanisms.

\begin{table}[bt]
    \centering
    \begin{tabular}{l|c|c}
    Agent & DoubleDQN & R2D2 \\
    \hline
    Input & $84\times84$ grayscale & $210\times 160$ RGB\\
    Action set & minimal per game: $3 \leq |\mathcal{A}| \leq 18$ & full: $|\mathcal{A}|=18$\\
    Reward & clipped & unclipped \\
    Neural net & feed-forward, 1.7M parameters & recurrent, 5.5M parameters \\
    Q-value head & regular & dueling \\
    Update & 1-step double Q-learning  & 5-step double Q-learning\\
    Optimiser & RMSProp without momentum & Adam with momentum $=0.9$ \\
    Batch size & $32$ & $32\times 80=2560$\\
    Replay, replay ratio & uniform, $8$ & prioritised (exponent $0.9$), $1$ \\
    Parallel actors & $1$ & $192$\\
    \hline
    Mean $\W$ per update & $\approx 9$\% & $\approx 6$\% \\
    \end{tabular}
    \vspace{1em}
    \caption{The two agent setups considered differ in a number of properties. Given both settings result in similar churn, none of them seem critical. See Appendix~\ref{app:dqn} and~\ref{app:r2d2} for details.}
    \label{tab:dqn-r2d2}
\end{table}

\subsection{Breadth of prevalence and non-causes}
\label{sec:range}
To judge the importance of the phenomenon, we need to establish whether it is specific to a narrow range of settings, or prevalent in a variety of domains, algorithmic variants, and hyper-parameters. 
It turns out that this is easy to do, because the effect is very much not a subtle one.
In fact, policy churn is present in two very different deep RL agents, namely DoubleDQN~\cite{ddqn} and (a variant of) R2D2~\cite{r2d2}, both widely used for training on Atari (see Appendix~\ref{app:dqn} and~\ref{app:r2d2} for agent and environment specifications). 
Despite the large differences between the algorithms summarised in Table~\ref{tab:dqn-r2d2}, the magnitude of policy change is surprisingly similar, indicating that it is unlikely that policy churn strongly depends on any of these specific choices.

The effect is also not specific to any environment: we measure similar magnitudes of policy change across a range of Atari games that vary in many dimensions, such as action space, reward scale and sparsity, deadliness, etc.
Furthermore, it is present in all stages of training (see Figures~\ref{fig:summary_churn} and~\ref{fig:typical-churn}). Unsurprisingly it is highest in early learning, but it remains high during training and even after evaluation performance converges or stabilises.
\footnote{Given the high churn in (preliminary) experiments on DM-Lab~\cite{beattie2016deepmind}, as shown in Figure~\ref{fig:dmlab}, we also consider it unlikely that the phenomenon is specific to the Atari setting.}

\subsection{Ablations}
\label{sec:ablations}

\paragraph{Redundant actions.}
A simple factor that could explain policy churn are redundant actions, i.e., when nominally different actions have the same outcome (in all, or a large fraction of state space). This property varies widely by environment (see also~\cite{nelson2021estimates}), but we observe similar levels of policy change across them (Figures~\ref{fig:summary_churn} and~\ref{fig:typical-churn}). Also, when exposing the full Atari action set in all games, creating explicit global redundancy, the churn magnitudes are not affected much (Figure~\ref{fig:r2d2-churn}). Appendix~\ref{app:redundant} looks at the fine-grained aspect of \emph{which} actions tend to be swapped for which others, and finds no structure easily related to the (known) equivalence relations (Figure~\ref{fig:confusion}).
In other words, most $\arg\max$ changes are \emph{not} happening between equivalent actions.

\begin{figure}[tb]
    \centerline{
    \includegraphics[width=0.6\textwidth]{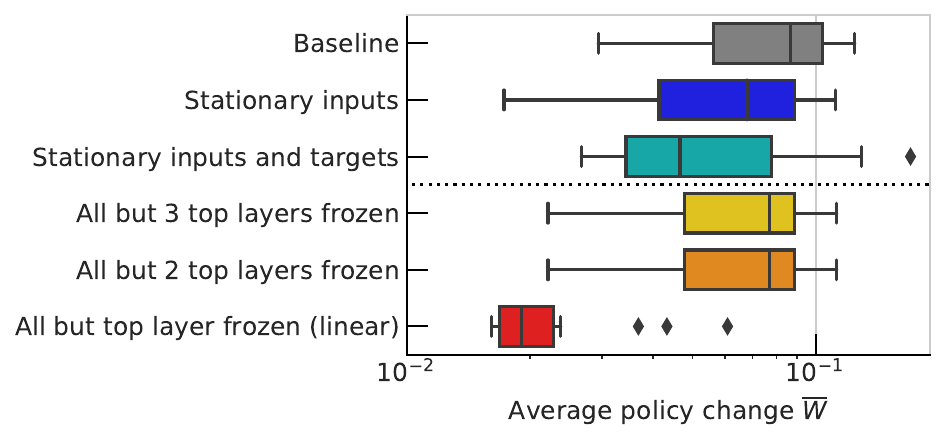}
    \hspace{0.01\textwidth}
    \includegraphics[width=0.38\textwidth]{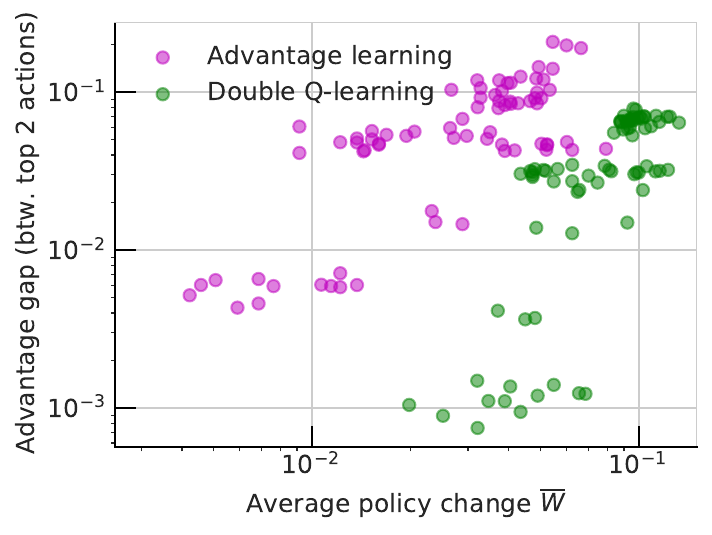}
    }
    \caption{{\bf Left:} Ablations in the ``forked'' DoubleDQN setting (5 games, 3 seeds each). Training progresses as normal until $50$M frames, after which either the data distribution is fixed (``stationary''), or part of the neural network is frozen; ``stationary targets'' denotes regression onto Monte-Carlo returns. The only large effect on policy change is when switching from deep learning to linear FA ({\color{red}\bf red}). For more detail see Figures~\ref{fig:forked-tandem}, \ref{fig:mc-returns}, and~\ref{fig:num-frozen-layers}. {\bf Right:} 
    Relation between policy change and advantage gaps. Different points correspond to different games, seeds and stages of learning. This shows that the action-gap-increasing algorithm ({\color{violet}\bf purple}) reduces policy change;
    and also how, given an algorithm, advantage gaps correlate with churn.
    For more detail see Figures~\ref{fig:advantage-churn} and~\ref{fig:advantage-e0}.
    }
    \label{fig:ablations}
\end{figure}

\paragraph{Small action gaps.} 
Another potential factor for large policy churn with greedy policies can be the interplay
between FA and small action gaps (difference between largest and second-largest action values): small approximation error can suffice for sub-optimal actions to overtake optimal ones. This hypothesis predicts that value learning methods inducing larger action gaps (e.g., Advantage Learning (AL,~\cite{action-gap}) which artificially lowers the values of sub-optimal actions) could reduce policy churn. Figure~\ref{fig:ablations} (right) shows that indeed policy churn is decreased substantially by AL, correlating with an increase of action gaps (which consistently grows under AL, see Figure~\ref{fig:advantage-churn}). Curiously, this does not seem to severely diminish the remaining policy churn's effectiveness for exploration: Figure~\ref{fig:advantage-e0} shows successful training of an AL-DQN with $\epsilon=0$.

\paragraph{Non-stationary state distribution.} 
Another explanatory hypothesis for the observed large magnitude of policy churn relates it to the non-stationary data distribution caused by 
the policy generating the data, which is evolving. Even when the agent converges to near-optimal performance, like in \textsc{Pong}, it may happen that the policy keeps changing on states where actions are inconsequential (where multiple actions have near-equal value), causing non-stationarity in the data distribution and thereby driving further policy churn. To test this, we utilize the ``forked tandem'' setting from \cite{tandem}, in which a high-performing policy is trained on the stationary data distribution generated by its initial snapshot (see below for more details). As can be seen in Figures~\ref{fig:ablations} (left) and~\ref{fig:forked-tandem}, the high level of policy churn is still preserved in this stationary-data regime, ruling out data non-stationarity as the main driver of the phenomenon.

\paragraph{Non-stationary targets.}
Temporal difference (TD) learning can give rise to another form of non-stationarity, as the bootstrap targets change at the pace of the target value function, an aspect that is preserved even in the forked tandem setting. 
To sidestep this, a simple control experiment uses the same setup but with Monte-Carlo returns as learning targets, turning it into pure policy evaluation via supervised regression (with noisy targets).
The results (Figures~\ref{fig:ablations}, left, and~\ref{fig:mc-returns}) show a similar level of churn to the Q-learning updates, indicating that the phenomenon is not specific to TD-based algorithms.

\paragraph{Decoupled acting and target networks.}
To assess \emph{how much} policy churn is beneficial for exploration, we devise the following setup:
A greedy policy based on the agent's target network drives behaviour, and we vary the frequency with which it is synchronized to the online network.
To avoid conflation with an adverse effect on learning stability, 
we keep the regular target network update frequency constant, while using an additional \emph{acting network}, periodically copied from the online network, for behaviour generation alone. Figure \ref{fig:acting-net} shows that an acting network updated more than every $\approx 1\,000$ gradient updates achieves most of the exploration benefits of acting with the online network, though in some games higher frequencies yield further benefit. This implies that the amount of policy change needed for exploration is much smaller than what is generated by DQN's learning process (otherwise the observed collapse would happen at much smaller update intervals).

\paragraph{Self-correction and the tandem effect.}
In \cite{tandem} a phenomenon dubbed the ``tandem effect'' was observed: the failure of a deep RL agent to adequately learn from the training data generated by a different instance of the same agent, highlighting the importance of \emph{self-correction} by interactively generated data from the policy being trained. One of their settings, the ``forked tandem'', starts with a copy of a high-performing policy and uses data sampled from its stationary distribution to continue training; even this apparently benign scenario leads to instability and potential collapse of the trained policy. 
Policy churn may provide a partial mechanistic explanation for the origin of the instability, showing that rapid policy change can be expected at all stages of training. The observation that the trained policy changes on a significant proportion of states at every update (performed on a negligibly small sample, a single minibatch of $32$ state transitions) supports the hypothesis from \cite{tandem} that erroneous extrapolation or over-generalization may play a key role in causing deviation and instability and producing the tandem effect in the absence of corrective training signal from self-generated data. Analogously to results in \cite{tandem} we observe that the magnitude of policy churn is highly correlated with the depth of the trained function approximator, further supporting this hypothesis; see Figures~\ref{fig:ablations} (left) and~\ref{fig:num-frozen-layers}, which show specifically how policy change drops dramatically in a linear FA regime.

\subsection{Detailed case study: \textsc{Catch}}
\label{sec:catch}

\begin{figure}[tb]
    \centering
    \includegraphics[width=\textwidth]{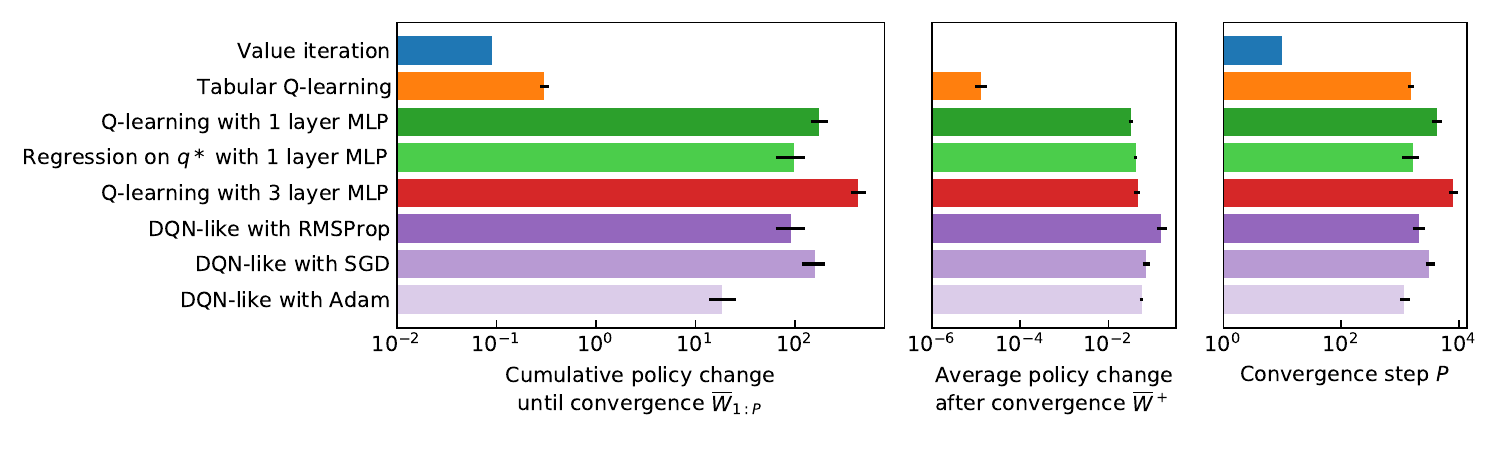}
    \caption{Ablations of what aspects drive policy change in \textsc{Catch}. Unless specified each FA variant uses the SGD optimiser. Here ``DQN-like'' is ``Q-learning with 3 layer MLP'' with replay and mini-batches. Error bars denote the interquartile range over $100$ seeds.}
    \label{fig:catch-case-study}
\end{figure}

\textsc{Catch} is a toy environment where the total number of states is small enough to be amenable to ground-truth dynamic programming approaches using the explicit matrix of transition probabilities, while providing an observation space that requires non-linear FA to represent the optimal value function.
We construct a spectrum of settings that all learn the optimal policy after some $P$ iterations. For each we measure cumulative policy change $\W_{0:P}$ and average change after convergence $\W^{+}$ (Eq.~\ref{eq:conv-churn}). Main results are shown in Figure~\ref{fig:catch-case-study}.
Sitting at one end of the spectrum, exact value iteration converges after $P=10$ steps with $\W_{0:P}= 0.09$ (because the initial policy is random, and the optimal policy has ties in most states).
The next simplest setting is tabular Q-learning with incremental updates (here and elsewhere, hyperparameters like the learning rate are tuned for fast convergence to optimal performance, see Appendix~\ref{app:catch} for details).
The next steps add in non-linear FA (shallow with $1$ hidden layers, then deep, with $3$).
At the other end of the spectrum is an approximation to DQN which includes experience replay, mini-batches, a deep neural network, and an advanced optimiser (RMSProp \cite{rmsprop}).
Figure~\ref{fig:catch-case-study} also shows intermediate variations such as supervised regression to $q^*$, and different optimisers (e.g., Adam~\cite{adam}); 
additional results in Appendix~\ref{app:more-catch}.
Overall, the experiments in this section show that, among the factors we considered, the presence of function approximation is by far the aspect that correlates the most with the occurrence of the phenomenon of policy churn.

\subsection{Mechanistic hypothesis}
\label{sec:hypothesis}

In an attempt to synthesise all the evidence presented so far, we propose that the observed policy churn is primarily a result of two components that need to be present jointly:
\begin{enumerate}[topsep=-0.2em,itemsep=0.2em,partopsep=0em,parsep=0em,leftmargin=0.7cm]
    \item Non-linear, global function approximation (such as deep neural networks), where each update can affect all states and all actions.
    \item A learning process with a high amount of noise. This could have multiple sources: stochastic optimisation (e.g. small batch sizes, large learning rates), noisy learning targets, non-stationary data (or targets, as with bootstrapping), an imbalanced data distribution (e.g., seeing some actions more often than others). 
\end{enumerate}
In other words, the rapid policy change dubbed churn is the symptom of high-variance updates to a global function approximator, and neither aspect in isolation is enough to promote it.\footnote{%
A compounding effect could be a \emph{mismatch} between the regression loss in value space that drives the learning process, and policy space, which is what matters for performance and for exploration (and is measured here). 
While supervised learning does not have this mismatch, it could conceivably have large ``policy change'' too. An in-depth treatment exceeds the scope for this paper, but preliminary results on MNIST (Appendix~\ref{app:mnist}) indicate \emph{no} large churn in the supervised setting.
}

\section{Where do we go from here?}
\label{sec:consequences}

\subsection{Learning at the edge of chaos}
\label{sec:chaos}
Deep learning has a well-known trade-off between speed of learning and stability that incentivises tuning the learning dynamics to be near the edge of chaos\footnote{As epitomised in the heuristic to tune the learning rate to the largest value that does not explode.}~\cite{cohen2021gradient}.
The presence of policy churn could enrich this picture in two ways for the case of deep RL. First, to the extent that rapid policy change helps drive exploration (Section~\ref{sec:exploration}), this is an additional incentive to keep learning dynamics sufficiently noisy.
Second, to the extent that value-based learning relies on self-correction~\cite{tandem}, that is, the actions to be corrected need to be picked by a greedy policy, this also encourages a large amount of policy change.
Circumstantial evidence for these is that value-based RL tends to require additional stabilisation mechanisms (e.g., target networks), less common in policy-based RL.

\subsection{Policy null-spaces}
\label{sec:null}

Policy change is necessary for policy improvement, but does not necessarily imply change in performance. 
We define the space of policies that have the same value  as a reference policy $\pi$ (in all states) as its \emph{null space} $\Null(\pi)$.
One way to quantify it is the \emph{diameter}
\[
|\Null(\pi)| := \max_{\pi', \pi'' \in \Null(\pi)} \W(\pi', \pi'')
\]
as the largest policy change possible between any pair of policies within it.
Typical scenarios with large null spaces have states in which actions have no effect (e.g., move actions while falling), multiple actions with the same effect (globally, or locally in part of state space), or multiple paths leading to the same outcome (e.g., up-then-left vs.~left-then-up).
These scenarios are common among the environments typically studied in deep RL.
Appreciating that policy null spaces can be large makes the policy churn phenomenon more palatable, helping us reconcile that agents changing their mind a million times about the best action (on average in each state) can have excellent performance.
A particularly interesting null space is the one around the optimal policy $\Null(\pi^*)$: when it is large, there are many ``safe'' ways to keep changing the policy (possibly by a lot) after converging to maximal performance, which is what we observe on, e.g., \textsc{Pong} (see Figure~\ref{fig:pong-late}), at least when $\epsilon > 0$.

\subsection{Off-policy corrections in the presence of churn}
\label{sec:off-policy}

Most types of off-policy correction are based on the gap between the data-generating behaviour policy $\mu$, and a target policy $\pi$. When doing multi-step updates, conservative methods truncate trajectories to bootstrap early compared to on-policy experience~\cite{kozuno2021revisiting}.
To obtain the benefit of multi-step value propagation, the average truncation length cannot be too short, and thus a silent assumption is that $\mu$ and $\pi$ do not differ too frequently.
Empirically however, multi-step back-ups without \emph{any} off-policy correction can be surprisingly effective \cite{rainbow}.
We can now re-interpret these findings through the lens of policy churn: as the greedy policy changes much faster than expected, even a slight latency (a few updates) between the parameters of the data-generating policy and the current target policy leads to massive truncation effect. If consecutive greedy policies are (approximately) within the null-space of each other, the benefits of an uncorrected multi-step update may outweigh its cost.
It also suggests the possibility of new off-policy algorithms that exploit the knowledge of the churn phenomenon, by truncating in a less aggressive fashion, motivated by the intuition that majority of rapidly changing policies lie within an (approximate) null space of each other (see Appendix~\ref{app:churn-offpolicy}).

\subsection{The social dynamics of research}
\label{sec:social}

If policy churn is indeed a hidden form of exploration, how did it come about? It seems unlikely that a useful mechanism emerges completely by chance. An intriguing possibility is that policy churn is the effect of a gradual process of \emph{natural selection}~\cite{darwin1859origin}. The hypothesis is that, if policy churn provides benefits, algorithms that display some level of it would be favoured over their counterparts in the inevitable engineering work surrounding the design of large-scale agents. In this view, RL practitioners play the role of ``nature'' exerting a selective pressure that shapes algorithms across time. This process could be completely \emph{unconscious} to the researchers involved: agents under-performing due to weak exploration would be discarded in favour of agents that explored better using some form of (hidden) policy churn. Over time, the multiple degrees of freedom of large-scale agents (hyper-parameters, network architecture, etc.) would be tuned to reflect just the right amount of churn: enough to help in exploration, but not too much to make the overall learning process unstable.

As appealing as the above hypothesis may be, as of now we do not have any evidence to support it. Still, it is worth considering, as it raises many interesting questions. Are there other hidden effects that have been selected for over the years? How many good design choices got discarded because they happened not to promote policy churn (or other similarly hidden effects)? Is the AI research community narrowly focused on a handful of design templates that happen to induce some ill-understood dynamics? This sort of question did not arise in the past, when agents were simple enough that we could keep track of their functioning at the finest level of detail. Now that deep RL agents have reached a certain level of complexity, any design choice may have a cascade effect whose consequences we do not anticipate. This creates the perfect environment for the sort of selective process described above. Acknowledging this possibility and being aware of it may be an important step toward unveiling hidden effects, and perhaps turning them into more purposeful design.

\section{Related work}
\label{sec:related}

A phenomenon in the literature that is related to policy churn, but not equivalent to it, is that of \emph{policy oscillation} or \emph{chattering}~\cite{gordon1995,bertsekas96neuro-dynamic,wagner2014policy}. Bertsekas and Tsitsiklis~\cite{bertsekas96neuro-dynamic} define the \emph{greedy region} of a function $q \in \Qspace$ as the set of functions in $\Qspace$ that induce the same greedy policy as $q$. Each greedy region has an associated ``fixed point'': the value function of the greedy policy induced by the functions in it. Conversely, every value function is the fixed point of a greedy region. It is well known that the only value function that belongs to its own greedy region is the optimal value function $q^*$~\cite{bertsekas96neuro-dynamic}. However, when function approximation is used, projecting value functions onto the FA space may create cycles that repeat indefinitely, a phenomenon known as policy oscillation. We do not think policy oscillation is a likely explanation for policy churn, because the approximators used in our experiments should have the capacity to approximate any value function to a reasonable level of accuracy. This is corroborated by Figures~\ref{fig:ablations} and~\ref{fig:catch-case-study} and Appendix~\ref{app:more-catch}, which show neural networks with more or wider layers exhibit \emph{more} churn.
We believe it is more instructive to think of each update of the approximator as moving the value function across the boundaries of greedy regions. As discussed, in many cases this has no effect on the agent's performance, since the policies associated with neighbouring greedy regions may belong to each other's null space (Section~\ref{sec:null}).

Another body of related work is the literature on stochastic gradient descent, and how a relatively high amount of stochasticity can be beneficial to optimisation by overcoming local optima and converging to flatter optima with better generalisation properties~\cite{keskar2016large,ruder2017overview,kleinberg2018alternative}; this is such a prominent effect that in some cases it is even beneficial to inject additional noise with Langevin dynamics \cite{welling2011bayesian}.
More loosely related are studies of learning in animals, exhibiting large drift in synaptic or representation space~\cite{mongillo2017,rule2021selfhealing,marks2020stimulus,Deitch2020drift,Schoonover2021},
as well as evidence for highly variable behaviour policies that can get consolidated through salient dopamine events~\cite{clopath2008tag}.

\section{Conclusions and Future Work}

\paragraph{Revisiting interpretations.}
Nine years after the introduction of DQN \cite{dqn-arxiv}, there are still phenomena in value-based deep RL that remain to be understood, with this paper putting the spotlight on one of them that lies at the intersection of learning and exploration dynamics.
In particular, we hope that an awareness of the churn phenomenon will make researchers revisit some good ideas that may have been prematurely disregarded or under-valued, either because their promised exploration effect was too entangled with learning dynamics and the resulting churn, or because they improved the stability of learning dynamics at the expense of reduced exploration that undid the overall gains.

\paragraph{Churn beyond value-based RL.}
An obvious follow-up question is whether policy churn is also an important effect beyond value-based algorithms.
We could imagine that actor-critic algorithms incur much less policy change, because stochastic policies change more smoothly, or because various penalty terms keep the updated policy from deviating too much from its precursor.
If that were the case, it would indicate that exploration in these agents differs as well, possibly with complementary advantages and disadvantages. Similarly, various instances of model-based RL~\cite{moerland2020model,muzero} may have less policy change, because the planning process could mitigate some of the function approximation effects. We leave these investigations to future work, but include one preliminary result (Figure~\ref{fig:actor-critic}) that seems to hint at churn being high in some actor-critic agents as well.

\paragraph{Explicit and controllable churn.}
If policy churn is indeed a valuable and non-trivial exploration mechanism, then it may be costly to deliberately abandon it, especially if its effect turns out to be complementary to simpler noise-based exploration mechanisms. 
Ideally we would want an explicit and controllable mechanism that produces the same kind of consistent, non-harmful exploration behaviour, but without such a tight entanglement with the learning dynamics.
The core benefit of such a division of labour would be that practitioners could study, change or tune the learning and exploration processes separately, without having to make arbitrary trade-offs (e.g., around stability, diversity, representations) that inevitably arise when they are considered in combination.

\begin{ack}
The ideas presented here were refined in discussion with numerous of our DeepMind colleagues. Will Dabney, Joseph Modayil and Matteo Hessel helped improve the paper with detailed feedback, and we thank David Silver, Diana Borsa, Miruna P\^{i}slar, Claudia Clopath, Vlad Mnih, Iurii Kemaev, Junhyuk Oh, Bilal Piot, Greg Farquhar, Dan Calian, Hado van Hasselt and Alex Pritzel for their input. We also thank the anonymous NeurIPS reviewers for their suggestions, as well as the many RLDM and ICML attendees whom we put our guesstimate survey question to.
\end{ack}

{
\bibliographystyle{abbrv}

\bibliography{main}

}

\clearpage

\appendix

\section{Additional Results and Ablations}
\label{app:more-ablations}

This appendix contains a number of figures that are already referenced from within the main paper.

\begin{figure}[b]
    \centering
    \includegraphics[width=\textwidth]{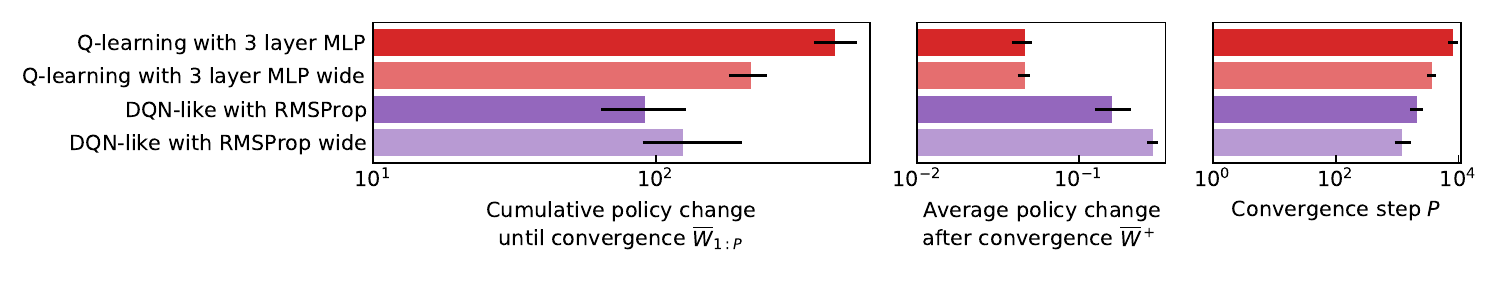}
    \caption{Ablation on the {\bf width} of the network in \textsc{Catch}. Here ``wide'' means the neural network has $200$ units per hidden layer instead of $50$. Increasing the width of the network increases policy change metrics $\W_{1:P}$ and $\W^+$ in the DQN-like variant whereas for the Q-learning with 3 layer MLP variant, it is the opposite.}
    \label{fig:catch-case-study-width}
\end{figure}

\begin{figure}[bt]
    \centering
    \includegraphics[width=\textwidth]{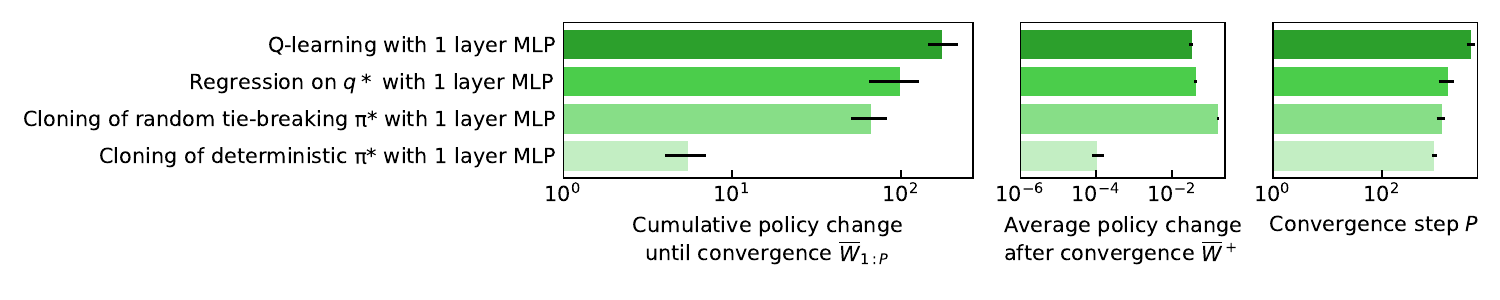}
    \caption{Additional supervised variants of \textsc{Catch}: behavioural cloning of $\pi^*$ with a cross-entropy loss and ground-truth targets. Note that the high $\W^+$ in the random tie-breaking variant is due to the many exact ties at the optimum; the deterministic variant has low $\W^+$.}
    \label{fig:catch-case-study-supervised}
\end{figure}

\begin{figure}[tb]
    \centering
    \includegraphics[width=\textwidth]{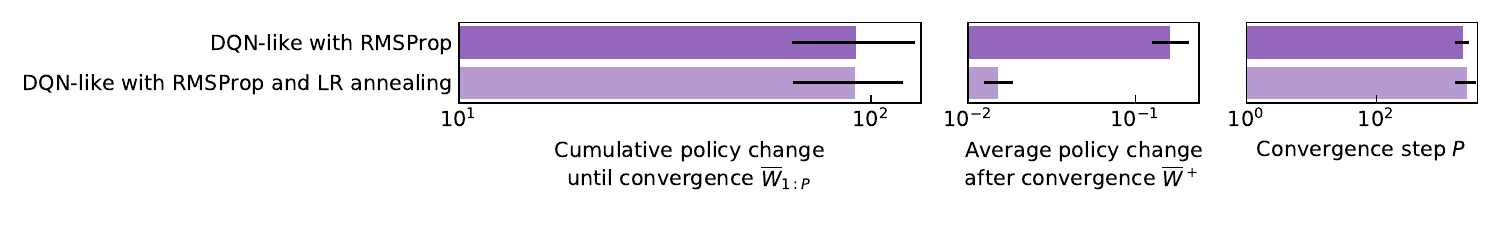}
    \caption{Variant on \textsc{Catch} where the learning rate is annealed from $10^{-3}$ to $10^{-4}$ over $10\,000$ steps. As expected the resulting average policy change after convergence is lower.}
    \label{fig:catch-case-study-lr-annealing}
\end{figure}

\subsection{Additional results on \textsc{Catch}}
\label{app:more-catch}

Complementing Figure~\ref{fig:catch-case-study} in Section~\ref{sec:catch} are Figures~\ref{fig:catch-case-study-width},~\ref{fig:catch-case-study-supervised}, and~\ref{fig:catch-case-study-lr-annealing} which show policy change for additional variants, in particular wider networks, behavioural cloning of $\pi^*$, and learning rate annealing.

\begin{figure}[p]
    \centering
    \includegraphics[width=0.9\textwidth]{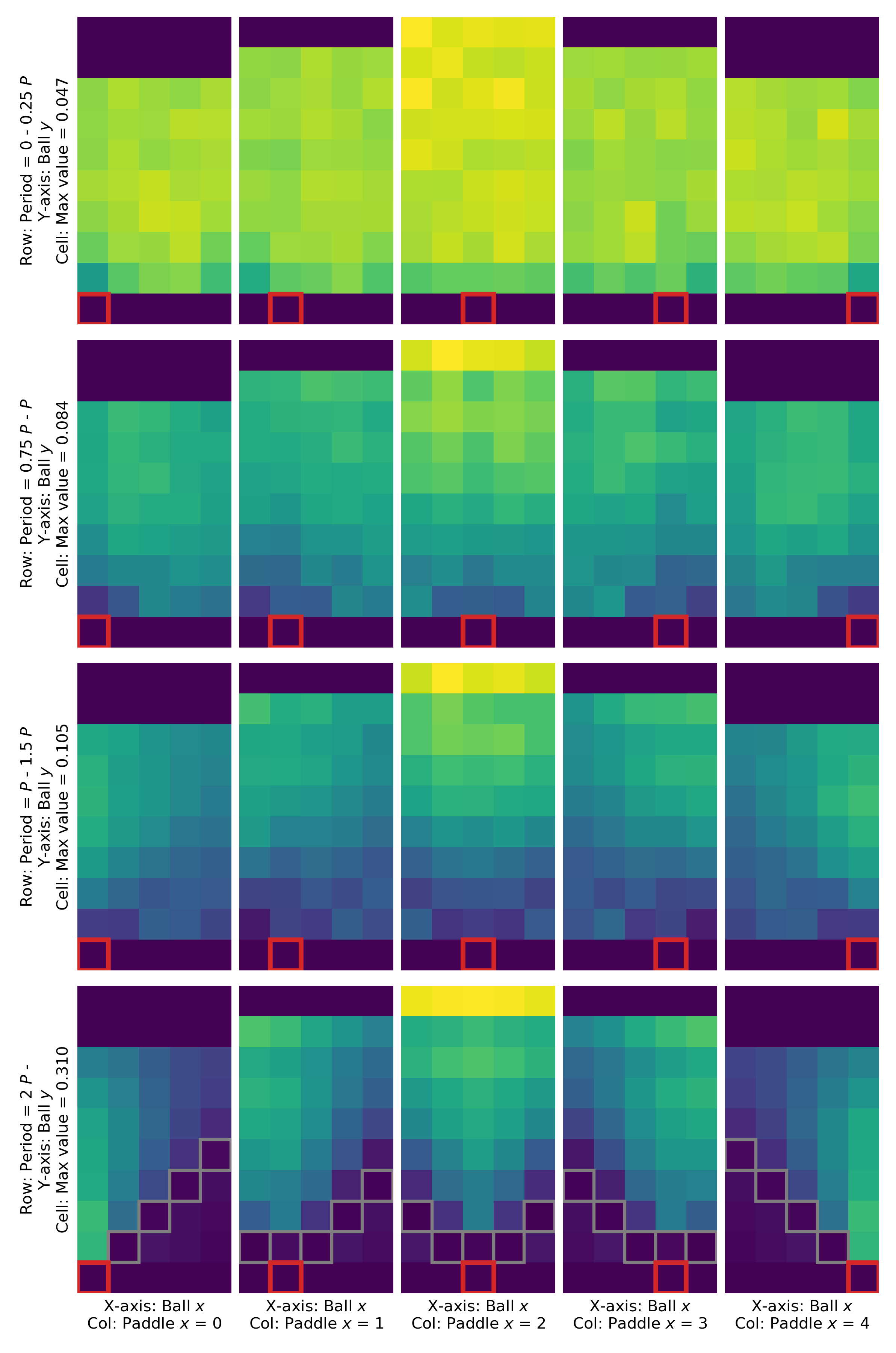}
    \caption{Policy change {\bf per state} averaged over different periods of training and seeds, on \textsc{Catch}. Each row of subplots represents a given period of training expressed in multiples of $P$ (performance convergence step), namely ``early'', ``pre-convergence'', ``post-convergence'' and ``late''. Each column corresponds to states where the paddle is at a particular $x$-coordinate, also highlighted by a red square. Each cell on a given figure represents the state corresponding to the $(x, y)$ position of the ball. The subplots in each row share the same scale, from $0$ to a $\max$ value indicated on the $y$-axis label. Cells highlighted by a grey square correspond to states where the action gap is non-zero for $q^*$, and we see that policy change is indeed lowest there, after convergence.
    See also Figure~\ref{fig:catch-grouped-by-gap}.
    }
    \label{fig:catch-policy-change-by-state}
\end{figure}

\begin{figure}[tbp]
    \centerline{
    \includegraphics[width=0.8\textwidth]{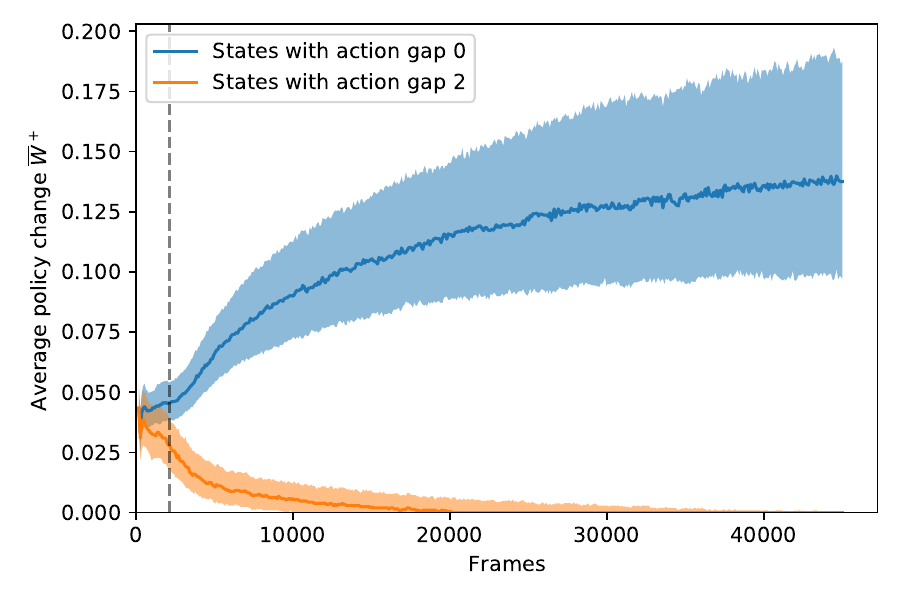}
    }
    \caption{Policy change across two groups of states in \textsc{Catch} with the DQN-like agent with RMSProp. The grey dotted line is the median convergence step $P$. Shaded areas denote the inter-quartile range over $1000$ seeds. This shows how late in training, and well after convergence, policy change concentrates in the null space (where action gaps are zero, blue) and critical actions (orange) are perturbed less and less often.
    See also Figure~\ref{fig:catch-policy-change-by-state}.
}
    \label{fig:catch-grouped-by-gap}
\end{figure}

\paragraph{Policy change per state.}
Figure~\ref{fig:catch-policy-change-by-state} shows the policy change per state averaged over different periods of training and $1\,000$ seeds for the ``DQN-like'' agent with RMSProp optimiser. See Appendix~\ref{app:catch} for exact hyper-parameters.
Note that episodes in \textsc{Catch} always start with the paddle in the centre. This means some states shown in the plots in Figure~\ref{fig:catch-policy-change-by-state} are not actually possible, in particular states corresponding to the dark top row of cells in all but the central column of plots. Another consequence is that starting states (corresponding to the top row of cells in the middle column of plots) have disproportionately more policy change. Indeed, in a version of the environment where the paddle is initialised randomly, this large relative difference in policy change disappears.
After convergence, in states where the action gap is high (states where the ball is diagonal from the paddle) there is little policy change as expected and most of the policy change happens in states corresponding to the ball higher up where the exact actions taken matter less; see also Figure~\ref{fig:catch-grouped-by-gap}. There is some conflation from the state distribution induced by the policy; everything else being equal, policy change is higher in states that are updated more often.
At the start of training and even a little while after convergence, for states where the paddle is on one of the sides (the first and last column of plots) and the ball is directly just above the paddle the relative amount of policy change is low. But well after convergence this flips and policy change is relatively high in these states. Presumably this is because early in training the agent has yet to learn that values for no-op action and the action that would move the paddle into the wall have the same effect.

\begin{figure}[ptb]
    \centering{
    \includegraphics[width=\textwidth]{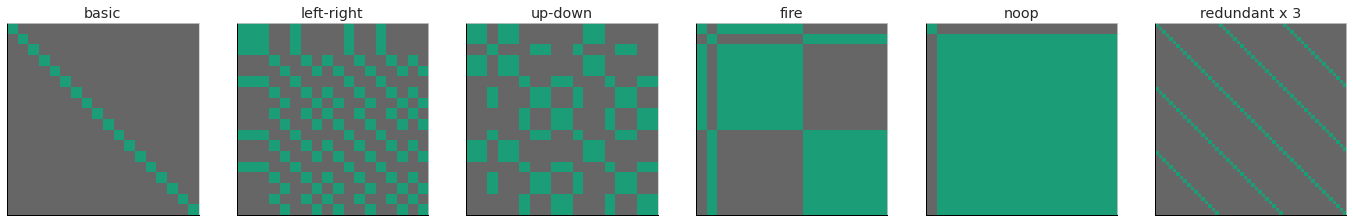}
    \includegraphics[width=\textwidth]{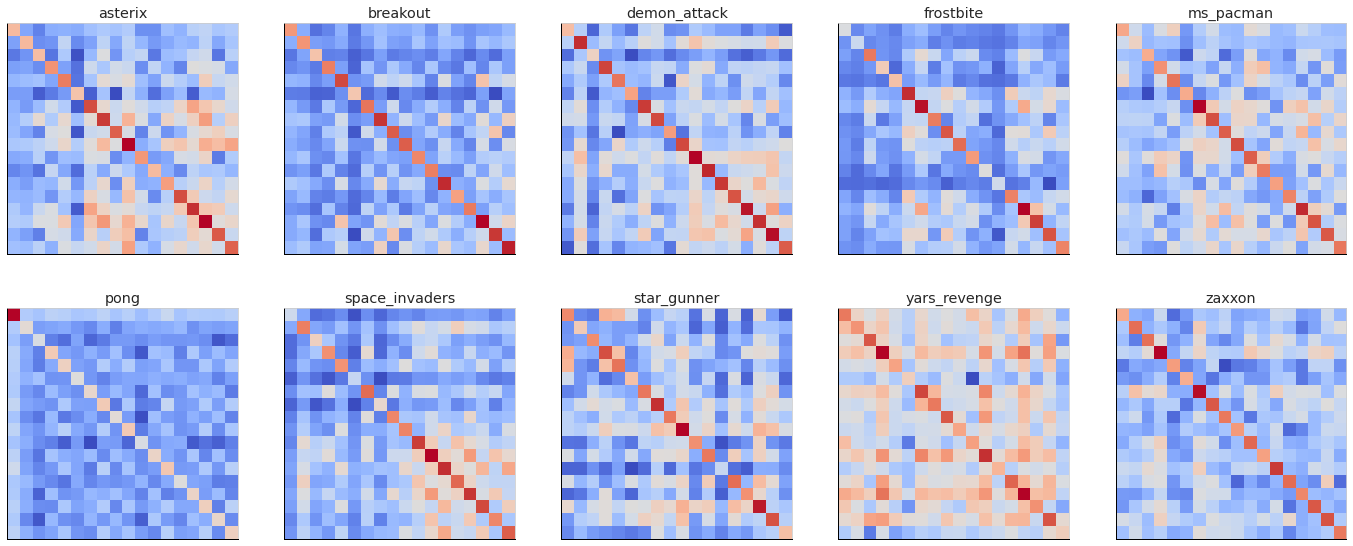}
    \includegraphics[width=\textwidth]{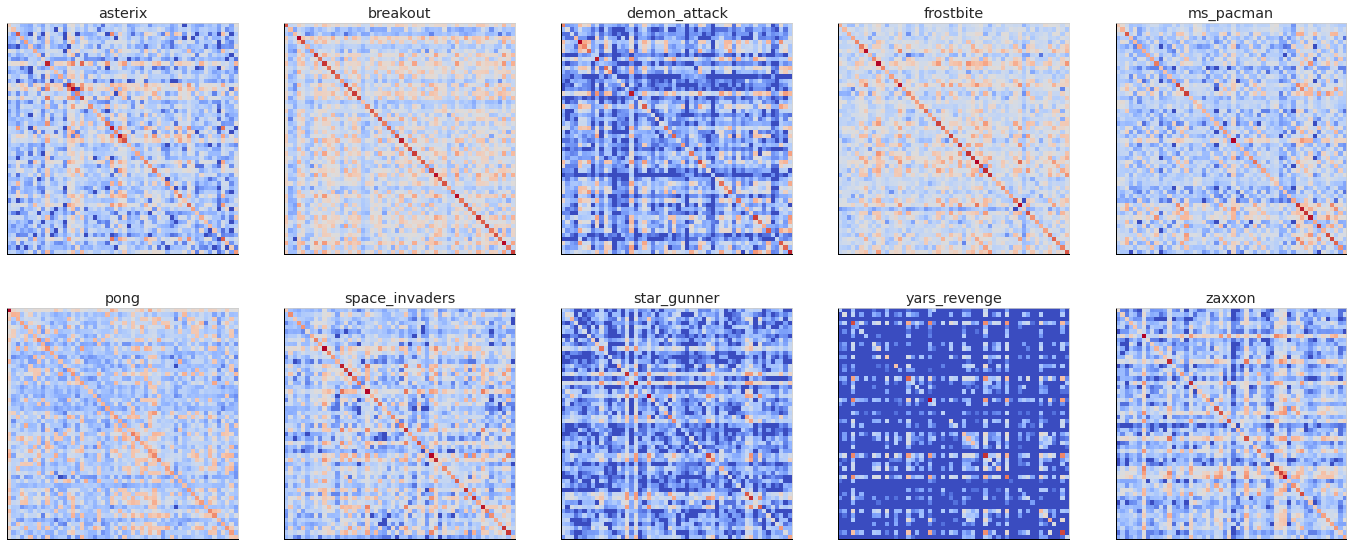}
    }
    \caption{Confusion matrices: between which actions do the $\arg\max$ switches happen?
    \textbf{Top row}: patterns that we could expect to see in games where all actions are distinct (``basic''), where only left-right movement matters (``left-right''), etc.
    \textbf{Middle rows}: empirical confusion statistics from an R2D2 experiment, warmer colors indicate higher likelihood (log-scaled).
    Note that some games have an effectively reduced action set; for example, in \textsc{Pong} only up/down/no-op matters, but this pattern (`up-down') does not show up in the switch statistics.
    \textbf{Bottom rows}: empirical confusion statistics in an ablation experiment where all actions were redundantly replicated three times (unbeknownst to the agent): here we would expect a pattern to emerge like in the top right (``redundant $\times 3$'') if the agent were to find out about the redundancy and only switch between these; but this does not happen.
    }
    \label{fig:confusion}
\end{figure}

\subsection{Redundant action spaces}
\label{app:redundant}

The DoubleDQN and R2D2 settings differ in the actions spaces used to act in the set of Atari games. As indicated in Table~\ref{tab:dqn-r2d2}, DoubleDQN always employs the minimal action set $3 \leq |\mathcal{A}| \leq 18$ (see subplot titles in Figure~\ref{fig:typical-churn}), while R2D2 always uses the full action set $|\mathcal{A}|=18$.
Adding to that, the experiments in Figure~\ref{fig:confusion} also include a ``redundant $\times 3$'' setting where the full action set is artificially replicated $3$ times ($|\mathcal{A}|=54$).

\subsection{Unlimited policy change in a two-armed bandit}
\label{app:2-arm-bandit}

One minimalist setting in which it is possible to obtain large (cumulative) policy change is incremental learning of similar Q-values using small step-sizes. For example, consider learning the two (tabular) Q-values of a two-armed bandit. Q-values are initialised near each other ($q_0(a_1)\approx q_0(a_2)$, and their true targets are also nearly identical ($q_P(a_1)\approx q_P(a_2)$, but far from initialisation, $q_0(\cdot) \ll q_P(\cdot)$.
With that set-up, a learning process that alternates between the actions to update can produce an $\arg\max$ switch on each update, because the last-updated Q-value will always be the larger one of the two.
And with the appropriate setting of step-sizes and initialisation, $P$ and thus $\W_{1:P}$ can be made arbitrarily large.

\subsection{High policy change in dynamic programming}
\label{app:adversarial-td}

Throughout the paper we treated policy change as an unexpected phenomenon. However, some amount of policy change is inherent to all RL algorithms. Value-based methods, in particular, are based on dynamic programming, which has at its core two operations: policy evaluation and policy improvement. Since by definition policy improvement involves change, it is fair to ask: how \emph{much} change is in fact expected? In other words: if we could isolate all other effects, like approximation and noise, how much policy change would still remain?

In Section~\ref{sec:catch} we already touched on this subject with the experiments on \textsc{Catch} using value iteration. In this section we revisit the question and try to provide a more definite answer to it. As it turns out, and perhaps not surprisingly, the answer to this question seems to be very domain dependent. The expected amount of policy change that is inherent to dynamic programming can vary significantly from one environment to the other. 

To illustrate this point, we now describe a simple policy evaluation setting that does not involve any approximation, incremental learning, or noise; and yet we see a large amount of policy change happening. Given the value function $q_\pi$ of a policy $\pi$, we compute the greedy policy $\pi'$ with respect to $q_\pi$, and monitor the changes in the greedy policy induced by the intermediate functions as we move from $q_\pi$ to $q_{\pi'}$. 

To describe our example precisely, we will need two concepts. First, we define the \emph{greedy operator} $g: \Qspace \rightarrow \Pi$ as
\begin{equation*}
    g(q) = \pi \text{ such that } \pi(s) = \mathrm{argmax}_a q(s,a), \text{ for all } s \in \mathcal{S},
\end{equation*}
where $q$ is an arbitrary function in $\Qspace$ and ties are broken in an arbitrary, but consistent, way. It will also be convenient to introduce the \emph{Bellman operator} of a policy $\pi$ as
\begin{equation*}
T_{\pi}q(s,a) = r(s,a) + \gamma \mathbb{E}_{S' \sim p(\cdot| s,a), A' \sim \pi(\cdot | S')} \left[ q(S', A') \right],
\end{equation*}
where $q \in \Qspace$, $r(s,a)$ is the expected reward following the execution of $a$ in $s$, $p(s'| s,a)$ is the probability of transitioning to state $s'$ given that action $a$ was executed in state $s$, and $\mathbb{E}[\cdot]$ is the expectation operator. It is well known that $\lim_{k \rightarrow \infty} T_{\pi}^k q  = q_{\pi}$ for any $q \in \Qspace$.

\begin{figure}[tb]
    \centering
    \includegraphics[scale=0.3]{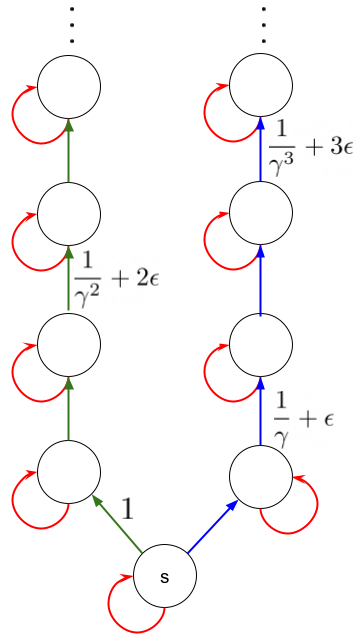}    
    \caption{Example of MDP in which a lot of policy change can happen at state $s$ during exact policy evaluation. States are represented as circles and actions are represented as arrows. Rewards are zero in all transitions except when marked otherwise in the diagram.}
    \label{fig:churny_mdp}
\end{figure}

Equipped with the concepts above, we can now present our example. Figure~\ref{fig:churny_mdp} shows an MDP composed of an arbitrary number of states structured as two chains. We are interested in monitoring how the policy will change in state $s$ as we do policy evaluation. Suppose that we start with a policy $\pi$ that selects action {\color{red} \bf red} everywhere. Clearly, $v_\pi(s) = 0$. The greedy policy $\pi' = g(q_\pi)$ will select actions associated with nonzero rewards whenever they are available; when they are not available, we will assume that the greedy operator $g$ will resolve the ties by always picking the {\color{OliveGreen} \bf green} or the {\color{blue} \bf blue} action over their {\color{red} \bf red} counterpart.

Starting from $q_{\pi}$, we will now monitor how much the greedy policy $g(T_{\pi'}^k q_{\pi})$ changes in $s$ with the sequence $k=1, 2, ...$, that is, as we move from $q_\pi$ to $q_{\pi'}$ as part of policy evaluation. For ease of exposition, we will use $\pi_k \equiv g(T_{\pi'}^k q_{\pi})$ to refer to the greedy policies along the way. Clearly, in the first step, when we change from $\pi$ to $\pi_1 = g(T_{\pi'}^1 q_{\pi})$, the policy changes in $s$ from {\color{red} \bf red} to {\color{OliveGreen} \bf green}. Now, in the second step, an easy calculation shows that the policy changes again, now from {\color{OliveGreen} \bf green} to {\color{blue} \bf blue}. If we keep doing this exercise, a simple pattern emerges: policies $\pi_k$ whose index $k$ is odd will pick action {\color{OliveGreen} \bf  green} in $s$, while their counterparts with an even index will instead select {\color{blue} \bf blue} on that state. This means that $W(\pi_k, \pi_{k+1} |s) = 1$  along the sequence of greedy policies $\pi_1, \pi_2, ...$. 

This deliberately simple example illustrates that the maximum possible amount of policy change can happen on a given state simply as an effect of policy evaluation. It is not difficult to construct examples in which a similar effect is observed throughout state space. 

In Section~\ref{sec:related} we discussed how the well-known policy oscillation effect may be responsible for part of the policy change when function approximation is used. The ``dynamic-programming effect'' discussed in this section happens in addition to that, regardless of function approximation. In general, we expect that policy change could be a result of both effects, plus other causes like the ones discussed in Section~\ref{sec:hypothesis} and Appendix~\ref{app:2-arm-bandit}. Given all the empirical evidence we have collected, we are reasonably confident that the causes discussed in Section~\ref{sec:hypothesis}---namely, global function approximation and noise---play a much more important role than the policy oscillation and dynamic programming effects in the setup studied.

\subsection{Churn-aware off-policy correction}
\label{app:churn-offpolicy}
Following up on Section~\ref{sec:off-policy}, this section spells out some concrete possibilities for forms of off-policy correction that take the churn phenomenon into account.
In a low-latency setting for example, it may be worth truncating traces when the noise of $\epsilon$-greedy leads to a low-advantage action getting executed, but not when the action discrepancy is purely due to churn ($\mu$ was acting greedily). Alternatively, we think it is plausible to make truncation decisions based on (relative) advantage gaps, effectively ignoring $\arg\max$ switches between actions of similar value.

\subsection{Relating churn to other game-specific properties}
\begin{figure}[tb]
    \centering
    \includegraphics[width=\textwidth]{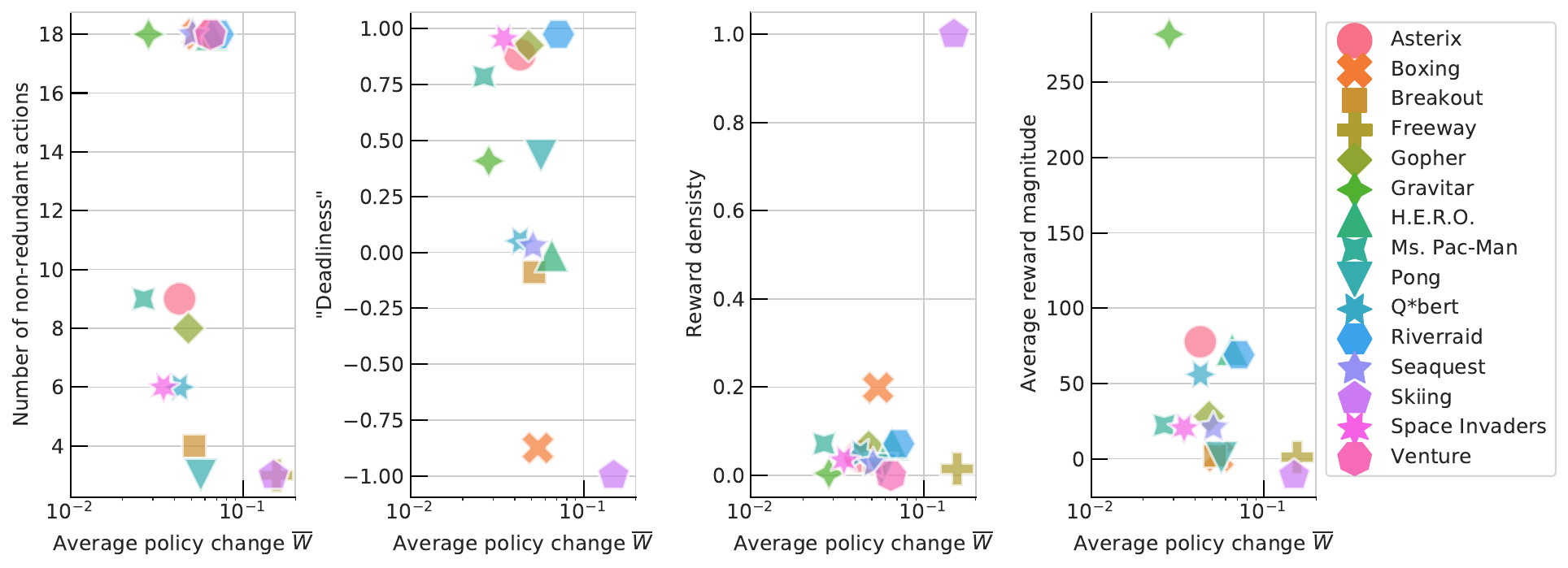}
    \caption{Relating average policy change $\W$ (across all seeds and time periods, in the R2D2 setup) to various game-specific properties. 
    The number of non-redundant actions refers to the minimal action set (used directly in DoubleDQN). ``Deadliness'' is the correlation between episodic return and episode length. Reward density is the fraction of transitions that produce a non-zero reward, and the average reward magnitude is the average return divided by the number of non-zero reward events.
    }
    \label{fig:bonus-scatters}
\end{figure}

Overall, we have not identified game-specific properties that are clearly predictive of the magnitude of policy change.
Figure~\ref{fig:bonus-scatters} provides a number of scatter plots for game-specific properties that we had considered as possibly having an influence.

\begin{figure}[tbp]
    \centerline{
    \includegraphics[width=0.5\textwidth]{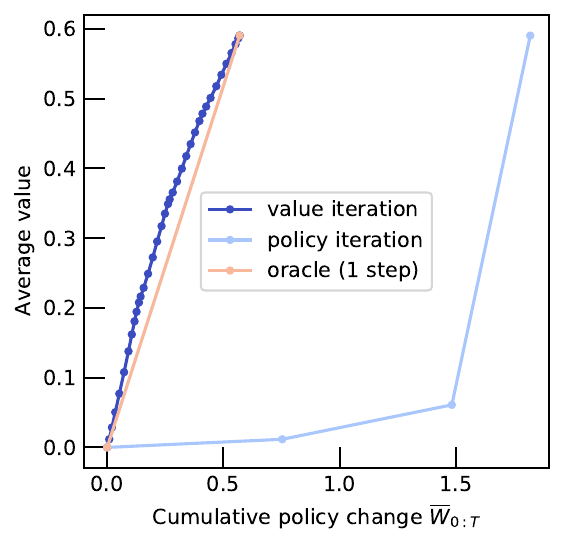}
    \includegraphics[width=0.5\textwidth]{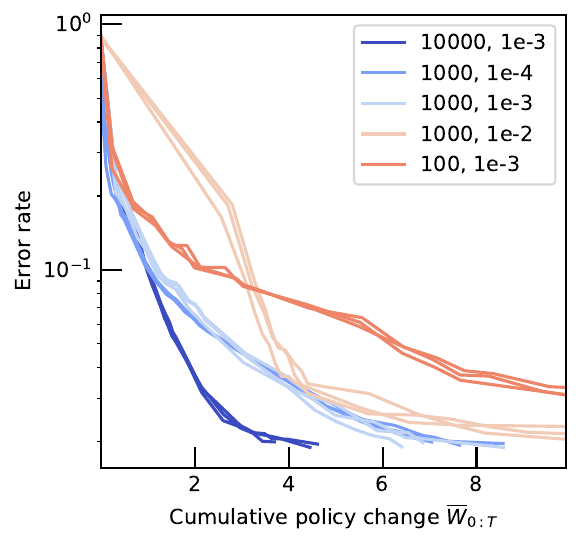}
    }
    \caption{Performance as a function of total accumulated policy change $\W_{1:T}$. 
{\bf Left:} Simple dynamic programming baselines in a tabular grid world. In this scenario, value iteration (blue) goes through $P=37$ steps until reaching $q^*$, but does not accumulate more policy change ($\W_{1:P}=0.57$) than an oracle that jumps from $q_0$ to $q^*$ (pink). Policy iteration does just $P=3$ steps but accumulates $\W_{1:P}=1.82$. Each iteration or update is shown as a dot. {\bf Right:} Supervised training of an MLP on MNIST with various hyper-parameter settings, listed as (batch size, learning rate) pairs in the legend. The multiple lines correspond to $3$ different random seeds for each setting.
Overall, vanilla MNIST training goes through a handful of label changes per input (on average) over the course of training.
}
    \label{fig:toy}
\end{figure}

\section{Experimental Details}
\label{app:experimental-details}

\subsection{DQN experiments}
\label{app:dqn}

\begin{figure}[ptb]
    \centering
    \includegraphics[width=\textwidth]{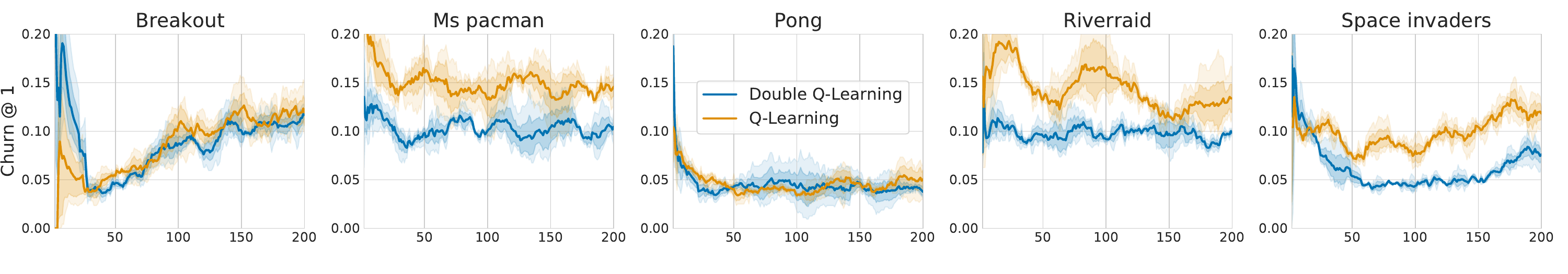}
    \includegraphics[width=\textwidth]{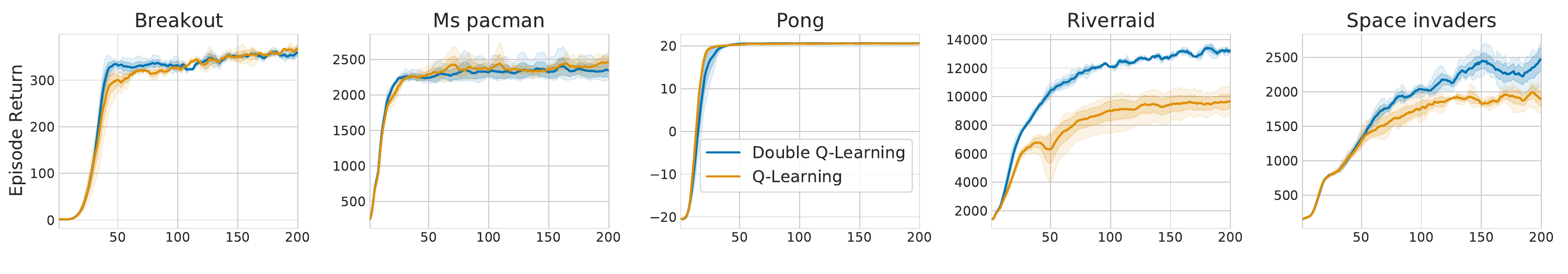}
    \caption{DQN versus DoubleDQN. Overall, DoubleDQN has somewhat better performance, while the level of policy change is a bit lower but not drastically different; in fact, the variation across games or across learning stages tends to be larger than the difference between algorithms.}
    \label{fig:dqn-ddqn}
\end{figure}

We chose to use double Q-learning with DQN (DoubleDQN, \cite{ddqn}) instead of vanilla DQN~\cite{dqn-arxiv,dqn} for all of our experiments,
as it is generally the more robust and better tuned of the two algorithms.
Apart from overall improved performance, for the purposes of this investigation there is little difference between the two, 
notably in terms of policy change, see Figure~\ref{fig:dqn-ddqn}.
We use an identical setting as the original DoubleDQN paper, including all hyper-parameters (which differ slightly from those in vanilla DQN). The main ones are listed in Table~\ref{tab:dqn-r2d2}, the remaining ones in Table~\ref{tab:hyper}. Our implementation is based on a slightly modified variant of the open-source DoubleDQN implementation in 
DQN Zoo~\cite{dqnzoo2020github}.

Our Atari investigations did not involve any hyper-parameter tuning. The modifications we did to existing settings for the exploration experiments (Section~\ref{sec:exploration}) are binary ablations:
\begin{itemize}
	\item Reducing $\epsilon=0.01$ to $\epsilon=0$ in the $\epsilon$-greedy behaviour policies
	\item Using the target network instead of the online network for acting.
\end{itemize}

The ``forked tandem’’ setup used in several ablations in Section~\ref{sec:ablations} follows \cite{tandem} and is based on their accompanying open-source implementation.\footnote{
\url{https://github.com/deepmind/deepmind-research/tree/master/tandem_dqn}}

Our Atari experiments are run with the same ALE variant of the Atari 2600 benchmark~\cite{ale} as in the original DQN and DoubleDQN works, using an action repeat of $4$, a zero discount on transitions involving a life loss, and the only source of stochasticity being a random number (uniformly between $0$ and $30$) of no-op actions applied at the beginning of each episode.  Unless stated otherwise, all these experiments are run with $3$ seeds for each configuration.

A lot of preliminary investigations used a small subset of Atari games (\textsc{Breakout}, \textsc{Pong}, \textsc{Ms. Pac-Man} and \textsc{Space Invaders}). For the final runs on $15$ games, we picked a representative subset of the $57$ Atari games, with a preference for games on which DoubleDQN can achieve a decent performance level.

\subsection{R2D2 experiments}
\label{app:r2d2}

The agent denoted as ``R2D2'' throughout the paper is a variant of the Recurrent Replay Distributed DQN architecture \cite{r2d2}. 
It comprises $192$ CPU-based actors 
concurrently generating experience and feeding it to a distributed experience replay buffer, 
and a single GPU-based learner randomly sampling mini-batches of experience sequences from replay
and performing updates of the recurrent value function by gradient descent. 
The value function is represented by a convolutional torso feeding into a linear layer, followed 
by a recurrent LSTM core, whose output is processed by a further linear layer before finally being 
output via a Dueling value head \cite{dueling}. The exact parameterization follows the 
slightly modified R2D2 presented in \cite{ezgreedy,return-scaling}, see Table \ref{tab:hyper} for a full
list of hyper-parameters.
It is trained using the Adam optimiser~\cite{adam} on a $5$-step Q-learning loss, using a periodically updated
target network for bootstrap target computation.
Replay sampling is performed using prioritized experience replay \cite{per} 
with priorities computed from sequences' TD errors following the scheme introduced in \cite{r2d2}. 
The agent uses a fixed replay ratio of $1$, i.e. the learner or actors are throttled dynamically if the average number
of times a sample gets replayed exceeds or falls below this value.
It also uses unclipped rewards and unclipped gradients, and an accompanying return-based normalisation, as in~\cite{return-scaling}.
Differently from those Atari RL agents following DQN~\cite{dqn}, our agent uses the raw 
$210 \times 160$ RGB frames as input to its value function (one at a time, without frame stacking),
though it still applies a max-pool operation over the most recent 2 frames to mitigate flickering
inherent to the Atari simulator. As in most past work, an action-repeat of $4$ is applied,  
episodes begin with a random number of no-op actions (up to $30$) being applied, and 
time-out after $108\,000$ frames (i.e. $30$ minutes of real-time game play).
The agent is implemented with JAX \cite{jax2018github}, uses the Haiku \cite{haiku2020github}, Optax 
\cite{rlax2020github}, Chex \cite{chex2020github}, and RLax \cite{optax2020github} libraries for neural networks, optimisation, testing, and RL losses, 
respectively, and Reverb \cite{Reverb} for distributed experience replay.

All our experiments ran for $40\,000$ learner updates. With a replay ratio of $1$, sequence length of $80$
(adjacent sequences overlapping by $40$ observations), a batch size of $32$, and an action-repeat of $4$ this corresponds 
to a training budget of $\approx 200$M environment frames
($\approx 100$ times fewer than the original R2D2). In wall-clock-time, one such experiment takes about $2$ hours.
All experiments are conducted across $15$ games, using $3$ seeds per game, unless stated otherwise.

\begin{table*}[tbp]
    \centering
    \begin{tabular}{l|c|c}
         Agent & DoubleDQN & R2D2 \\
        \hline
         Convolutional torso channels & $32, 64, 64$ & $32, 64, 128, 128$ \\
         Convolutional torso kernel sizes & $8, 4, 3$ & $7, 5, 5, 3$ \\
         Convolutional torso strides & $4, 2, 1$ & $4, 2, 2, 1$ \\
         Pre-LSTM linear layer units & N/A & $512$ \\
         LSTM hidden units & N/A & $512$ \\
         Post-LSTM linear layer units & N/A & $256$ \\
         Value head units & 512 & Dueling $2 \times 256$ \\
         Action repeats & $4$ & $4$ \\
         Actor parameter update interval & $4$ steps & $400$ steps \\
         $\epsilon$ for $\epsilon$-greedy policy & annealed from $1$ to $0.01$ & fixed $0.01$ \\
         Replay sequence length &$1$ & $80$ \\ 
         Replay buffer size & $10^6$ & $4\times 10^6$ observations \\
         Priority exponent & N/A & $0.9$ \\
         Importance sampling exponent & N/A & $0.6$ \\ 
         Discount $\gamma$  &$0.99$ & $0.997$ \\
         Target network update interval & $120\,000$ frames ($7\,500$ updates) & $400$ updates \\
         Gradient clipping & $\frac{1}{32}$ & N/A\\
         Normalisation & N/A & Return-based~\cite{return-scaling}\\
         Optimiser \& settings & RMSProp \cite{rmsprop} & Adam \cite{adam}, \\
         & learning rate $\eta = 2.5 \times 10^{-4}$, & learning rate $\eta = 2\times10^{-4}$,  \\
         & decay $= 0.95$, $\epsilon = 10^{-6}$ &  $\beta_1=0.9$, $\beta_2=0.999$, $\epsilon=10^{-6}$ \\

    \end{tabular}
    \caption{Atari agent hyper-parameter values (in addition to those in Table~\ref{tab:dqn-r2d2}). These follow~\cite{ddqn} and \cite{return-scaling}, respectively.}
    \label{tab:hyper}
\end{table*}

\subsection{\textsc{Catch} experiments}
\label{app:catch}

For \textsc{Catch}~\cite{bsuite} experiments, Table~\ref{tab:catch-variant-settings} lists the hyper-parameters for each of the variants specified in Figure~\ref{fig:catch-case-study}.
For each variant, seeds that did not converge after $5\,000$ episodes of training were filtered out. In practice, all seeds for all variants in the table converged. For all \textsc{Catch} experiments convergence is defined as when the greedy policy achieves the maximum score for $100$ evaluation episodes. Convergence is periodically tested every $100$ training episodes.

\begin{table}[tbp]
    \centering
    \begin{tabular}{l|l|c}
    \hline
    Value iteration & & \\
    \hline
    Tabular Q-learning & Learning rate & $0.1$ \\
                       & Batch size    & $1$ \\
    \hline
    Q-learning with 1 layer MLP & Learning rate    & $0.1$ \\
                                & Batch size       & $1$ \\
                                & Optimiser        & SGD \\
                                & \# hidden layers & 1 \\
    \hline
    Regression on $q^*$ with 1 layer MLP & Learning rate    & $0.1$ \\
                                        & Batch size       & $1$ \\
                                        & Optimiser        & SGD \\
                                        & \# hidden layers & 1 \\
    \hline
    Q-learning with 3 layer MLP  & Learning rate    & $0.1$ \\
                                 & Batch size       & $1$ \\
                                 & Optimiser        & SGD \\
                                 & \# hidden layers & 3 \\
    \hline
    DQN-like with RMSProp & Learning rate           & $0.001$ \\
                          & Batch size              & $32$ \\
                          & Optimiser               & RMSProp \\
                          & Optimiser $\varepsilon$ & $10^{-5}$\\
                          & Replay capacity         & $1\,000$ \\
                          & \# hidden layers        & 3 \\
    \hline
    DQN-like with SGD     & Learning rate    & $0.01$ \\
                          & Batch size       & $32$ \\
                          & Optimiser        & SGD \\
                          & Replay capacity  & $1\,000$ \\
                          & \# hidden layers & 3 \\
    \hline
    DQN-like with Adam    & Learning rate           & $0.001$ \\
                          & Batch size              & $32$ \\
                          & Optimiser               & Adam \\
                          & Optimiser $\varepsilon$ & $10^{-8}$\\
                          & Replay capacity         & $1\,000$ \\
                          & \# hidden layers        & 3 \\
    \hline
    (Common hyper-parameters) & Exploration $\epsilon$   & $0.1$ \\
                              & \# units per hidden layer & $25$ \\
    \hline
    
    \end{tabular}

    \caption{\textsc{Catch} case study variant settings. These are the relevant settings for the variants used to generate Figure~\ref{fig:catch-case-study}.}
    \label{tab:catch-variant-settings}
\end{table}

\subsection{Dynamic programming}
\label{app:toy-mdp}
To measure policy change of dynamic programming in a tabular MDP, we exploit the knowledge of the exact transition dynamics, encoded via a matrix $\mathbf{T}$ to compute value or policy iteration updates that do not involve sampling or interactions. Values are initialised at $0$, and for the purposes of measuring policy change, all  $\arg\max$ actions whose Q-values are exactly tied also share equal probability mass.
As example domain we use a $16\times16$ Gridworld with $4$-room structure, initial state in one corner, goal state in opposing corner and $\gamma=0.97$.
Figure~\ref{fig:toy} (left) shows the amounts of policy change accumulated in such a process.

\subsection{MNIST experiments}
\label{app:mnist}
For a simple initial supervised learning experiment, we used an off-the-shelf neural network training setup on MNIST.
Thus we used a $3$-layer MLP with $300$ and $100$ hidden units, ReLU non-linearities, a softmax output, cross-entropy loss and the Adam optimiser~\cite{adam}.
Policy change is measured on the softmax probability outputs of the classification network, with equal weight on all samples of the test set. It is accumulated across all $P$ gradient updates. Our experiments are stopped when reaching $2\%$ training error, which happens after $P\approx 1\,000-10\,000$ updates.  Figure~\ref{fig:toy} (right) shows the results.

\begin{figure}[ptb]
    \centerline{
    \includegraphics[width=\textwidth]{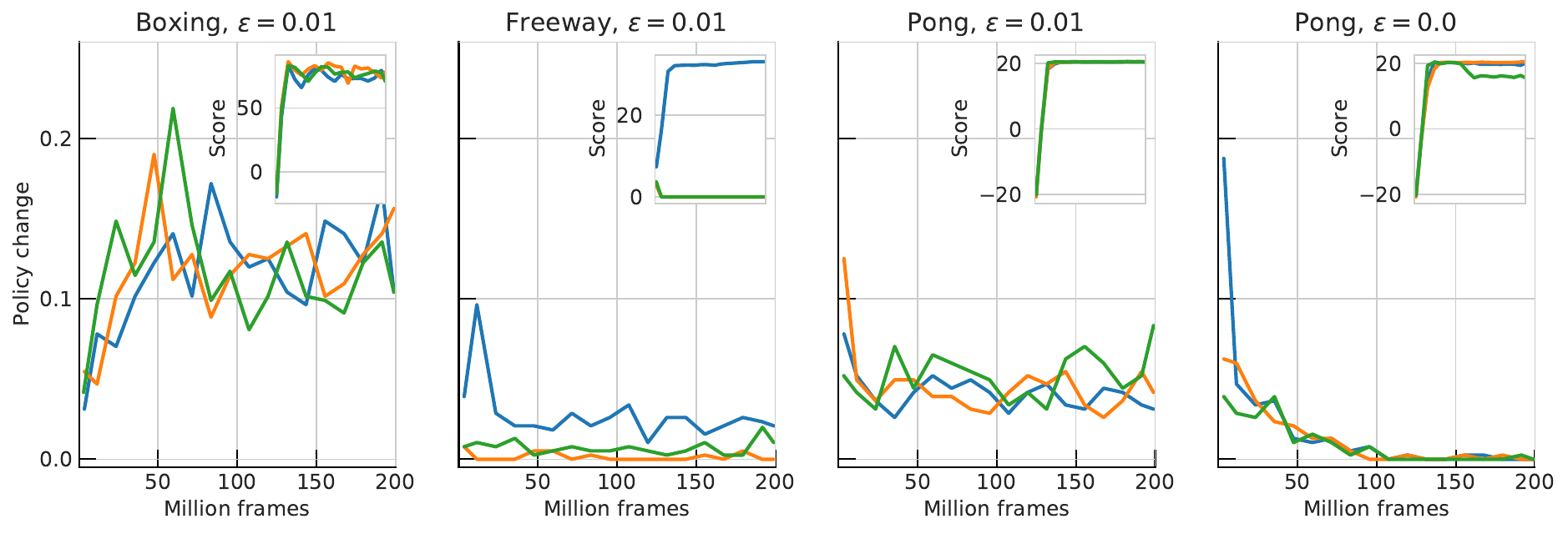}
    }
    \caption{Policy change on plateaus. We observe a high amount of policy change (per single update, i.e., $\W(\pi_{t}, \pi_{t+1})$) even in periods where overall policy performance is flat (see performance curves on inset plots). Each curve corresponds to a single run (seed) and is smoothed over 10M frames. 
    An interesting effect to highlight is in \textsc{Freeway}, where one seed ({\color{blue}blue}) converges to high performance and the other two seeds collapse to zero performance, and the ``broken'' runs also have much lower churn.
    The right-most figure shows (on \textsc{Pong}) that converged performance, together with $\epsilon=0$ leads to policy change that eventually \emph{does} seem to decay. 
    }
    \label{fig:pong-late}
\end{figure}

\begin{figure}[ptb]
    \centering
    \includegraphics[width=\textwidth]{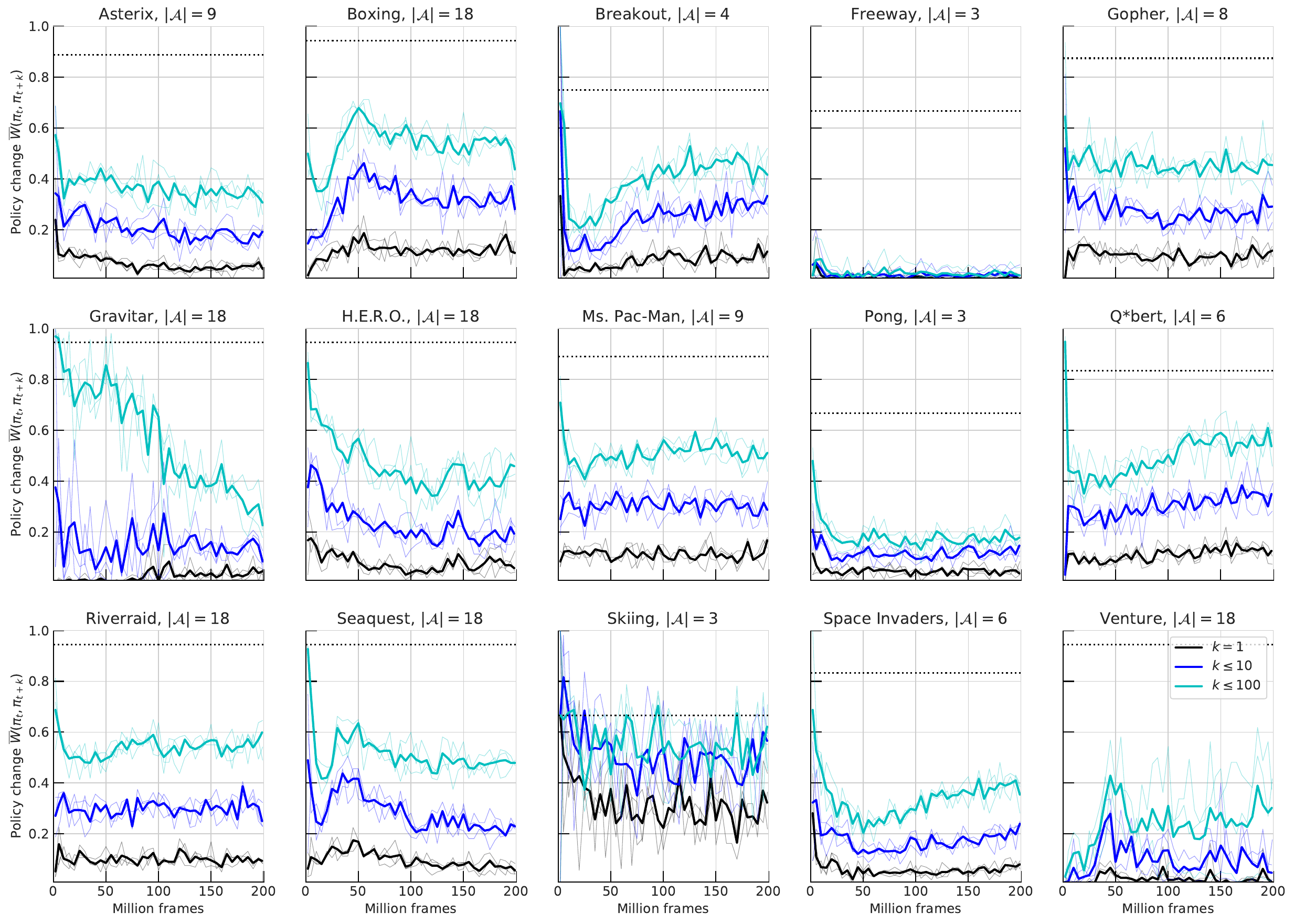}
    \caption{Average policy change $\W$ as a function of training stage in DoubleDQN, across 15 Atari games. 
    Often, but not always, policy change is larger in early learning. 
    Different colours show different interval sizes $k$ across which $\W(\pi_t,\pi_{t+k})$ is measured. 
    In some scenarios these show a more cumulative effect (e.g., \textsc{Gravitar}), in others the change between the very next policy ($k=1$) is almost as large to the change after $k=100$ updates, as in \textsc{Skiing}.
    Dotted lines indicate what ``maximal'' policy change would look like for a given action space size $|\mathcal{A}|$, i.e., if $\arg\max$ actions were completely random. Thin lines are individual seeds ($3$), thick lines their average.
    See Figure~\ref{fig:pong-late} for a detailed look at the low levels of policy change after convergence as in \textsc{Pong} or \textsc{Freeway}.}
    \label{fig:typical-churn}
\end{figure}

\begin{figure}[ptb]
    \centering
    \includegraphics[width=\textwidth]{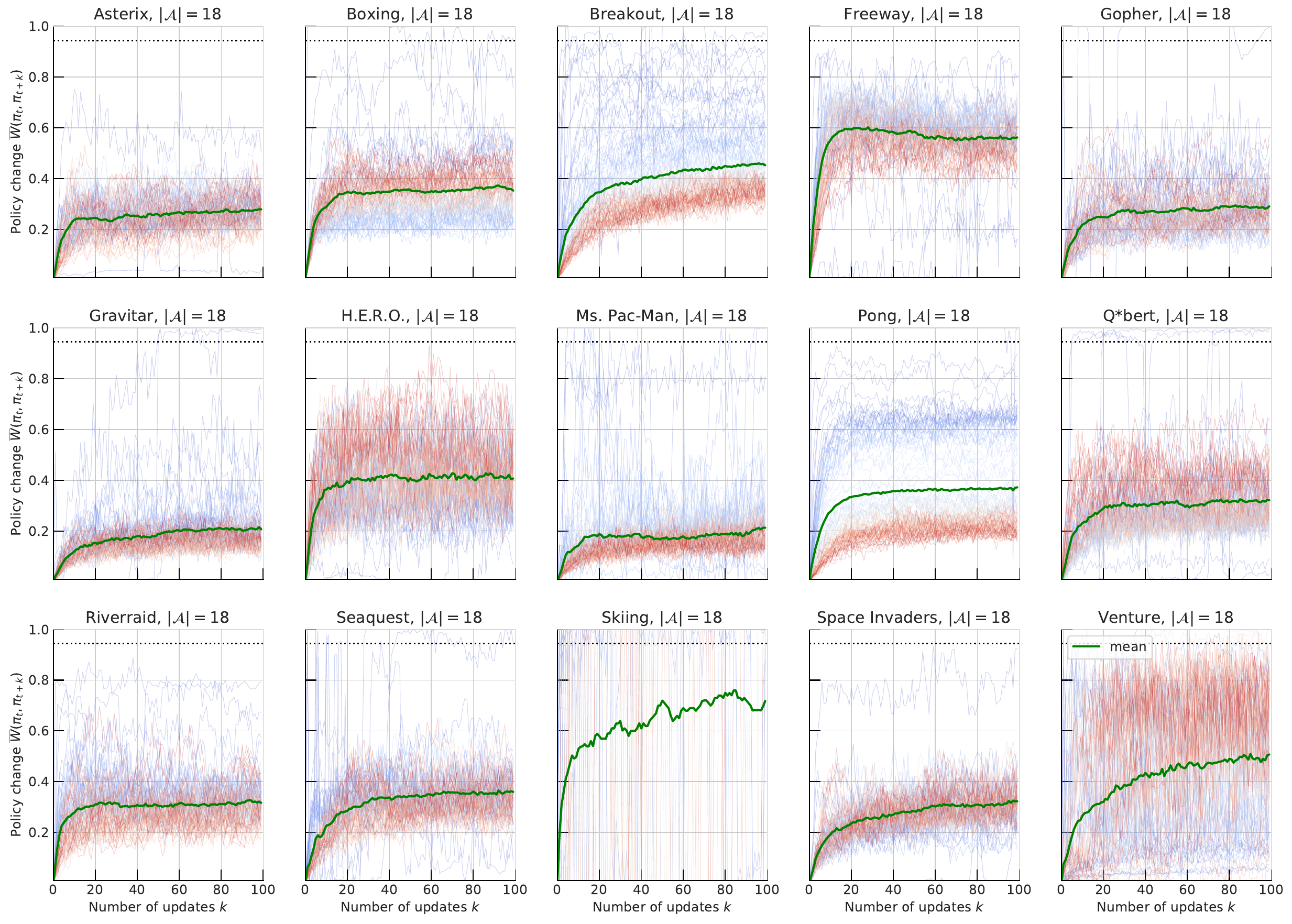}
    \caption{Average policy change $\W(\pi_t, \pi_{t+k})$ as a function of the number of in-between updates $k$. In contrast to Figure~\ref{fig:typical-churn}, these results are from the R2D2 agent (always using the full action set of $|\mathcal{A}|=18$).
    Thick green lines show the average across training, while thin lines show snapshots from different points in training, with cooler and warmer colors denoting early and late stages of training respectively. 
    We can see how policy change quickly rises and then saturates, generally between $20\%$ and $60\%$. This means that, compared to $\pi_t$, the policy $\pi_{t+100}$ generally does not differ much more than $\pi_{t+20}$. 
    This is consistent with the hypothesis that policy churn only affects a subset of states.
    An outlier here are the \textsc{Skiing} results, where the observed fraction of $\arg\max$ switches (in a minibatch of $32\times80$ states) is always either $0$ or $1$: this seems to indicate that the Q-values have essentially no state-dependence (note that performance also does not take off in this game, see Figure~\ref{fig:r2d2-performance}).
    }
    \label{fig:r2d2-churn}
\end{figure}

\begin{figure}[ptb]
    \centering
    \includegraphics[width=\textwidth]{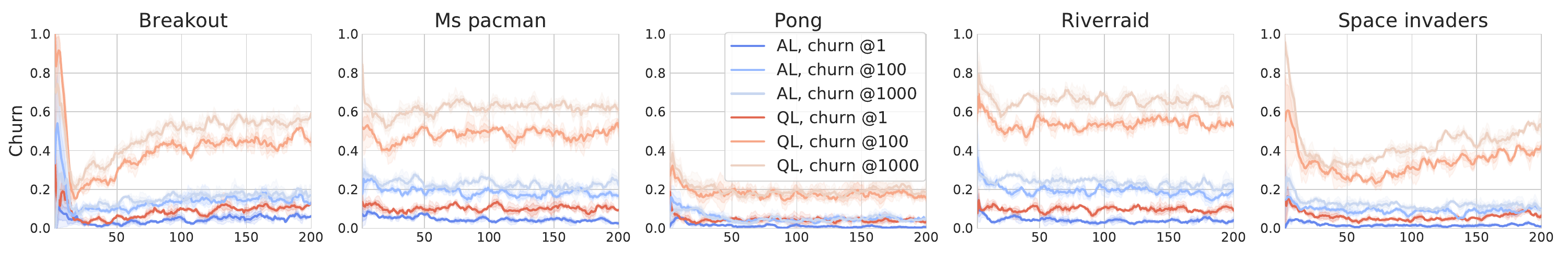}
    \includegraphics[width=\textwidth]{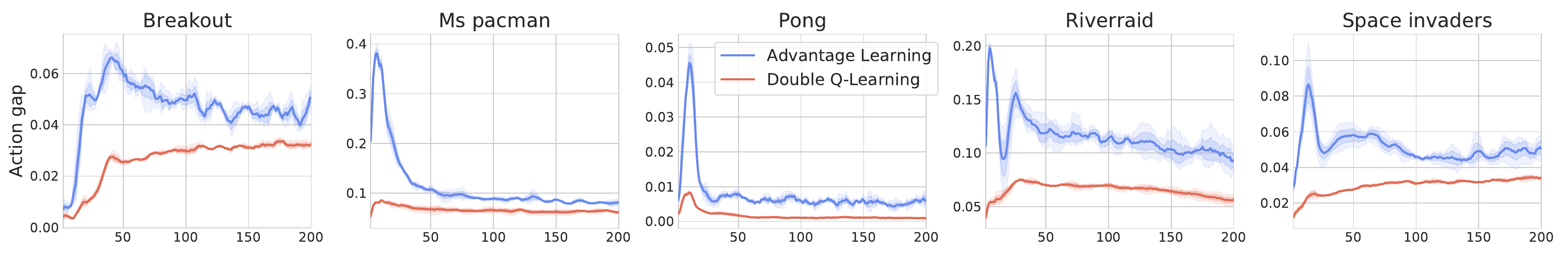}
    \caption{Double Q-Learning (``QL'', red) versus action-gap-increasing Advantage Learning (``AL'', blue), on a set of 5 games. 
    {\bf Top:} policy change, where ``@$k$'' denotes the interval in $\W(\pi_{t}, \pi_{t+k})$.
    {\bf Bottom:} corresponding action gaps.
    This provides time-series detail to Figure~\ref{fig:ablations} (right).
    }
    \label{fig:advantage-churn}
\end{figure}

\begin{figure}[ptb]
    \centering
    \includegraphics[width=\textwidth]{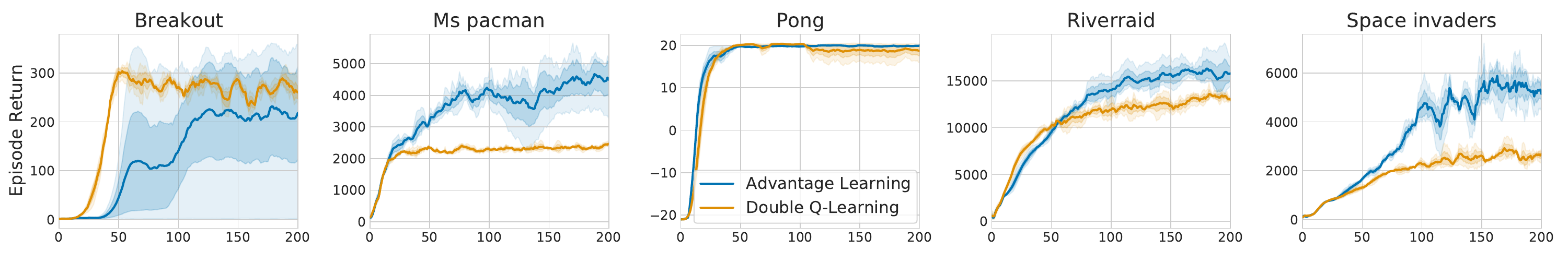}
    \caption{Performance results of Double Q-Learning and Advantage Learning, as in Figure~\ref{fig:advantage-churn}, but in the $\epsilon=0$ setting.
    Despite reduced churn, Advantage Learning is the higher-performing algorithm, indicating that not the full amount of DoubleDQN's observed policy change is needed for performance, even in the absence of other forms of exploration. This matches the insights in Figure~\ref{fig:acting-net}.
    }
    \label{fig:advantage-e0}
\end{figure}

\begin{figure}[ptb]
    \centering
    \includegraphics[width=\textwidth]{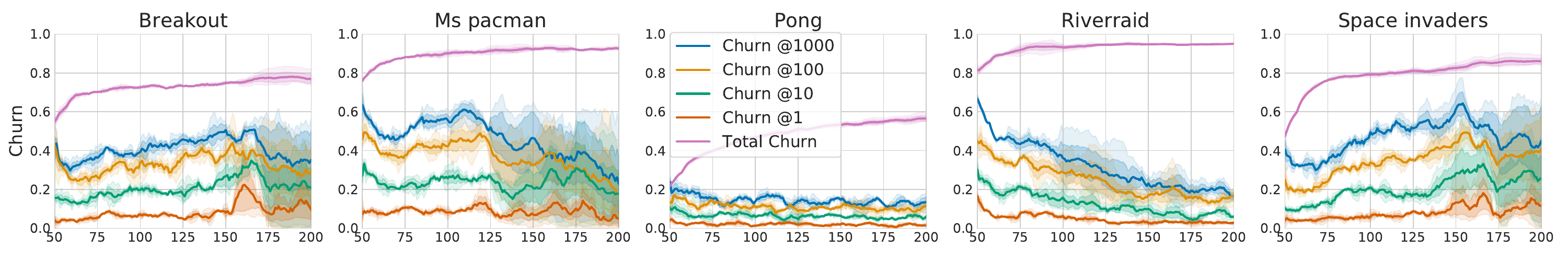}
    \includegraphics[width=\textwidth]{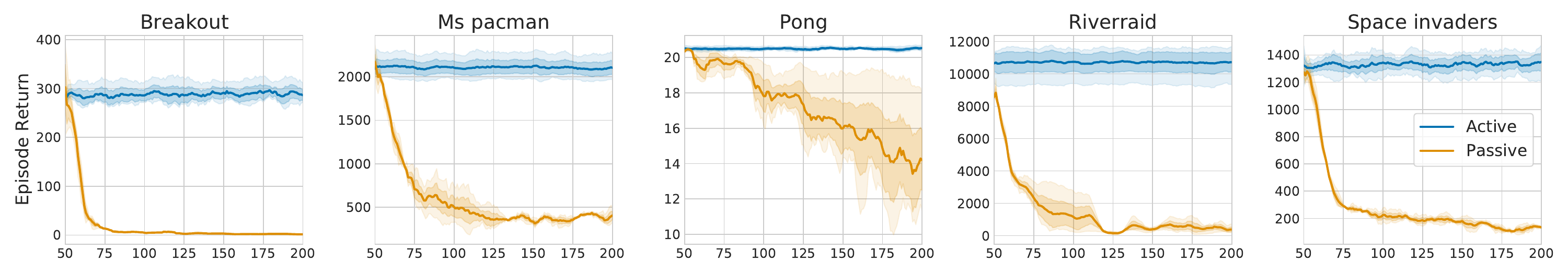}
    \caption{Stationary data in the ``forked tandem'' setting: after $50$M frames (start of $x$-axis), a passive learner is forked off, which means that it does not influence behaviour anymore (and cannot self-correct). It receives a data stream from a fixed, frozen policy network.
    {\bf Top:} active vs.~passive performance. 
    {\bf Bottom:} policy change. the purple curve (``total churn'') denotes the difference between the active (frozen) policy and the current policy of the passive (but learning) network.
    }
    \label{fig:forked-tandem}
\end{figure}

\begin{figure}[ptb]
    \centering
    \includegraphics[width=\textwidth]{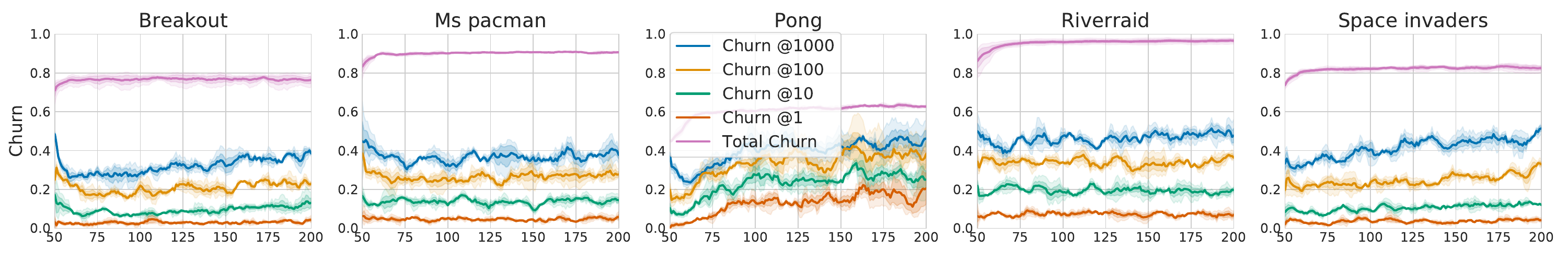}
    \caption{Stationary data \emph{and targets}. Setup as in Figure~\ref{fig:forked-tandem}, but instead of Q-learning bootstrap targets, stationary regression targets are constructed from Monte-Carlo returns.}
    \label{fig:mc-returns}
\end{figure}

\begin{figure}[ptb]
    \centering
    \includegraphics[width=\textwidth]{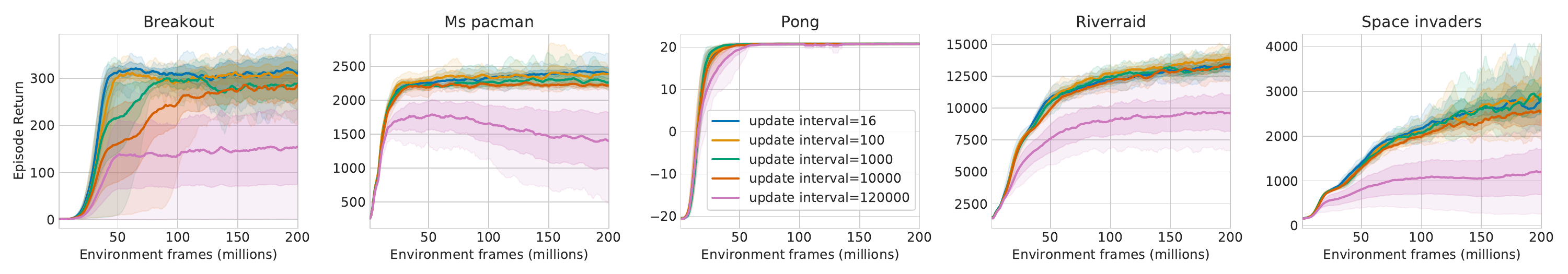}
    \caption{Ablation experiment with a separate copy of the Q-network used exclusively for acting; this network is a periodic copy of the online (learning) network, just like the target network, but updated at a different frequency. ``Interval~$=16$’’ corresponds to the DoubleDQN baseline, while  ``Interval~$=120\,000$’’ corresponds to the ``act with target network’’ of Section~\ref{sec:exploration} and Figure~\ref{fig:zero-eps} (denoted ``no churn’’ there).
We find again (cf. Figure~\ref{fig:advantage-e0}) that the full empirical magnitude of policy change in DoubleDQN is not needed for exploration: reducing the number of different greedy policies used for acting by a factor $100-1\,000$ still retains a very similar exploration effect.
}
    \label{fig:acting-net}
\end{figure}

\begin{figure}[ptb]
    \centering
    \includegraphics[width=\textwidth]{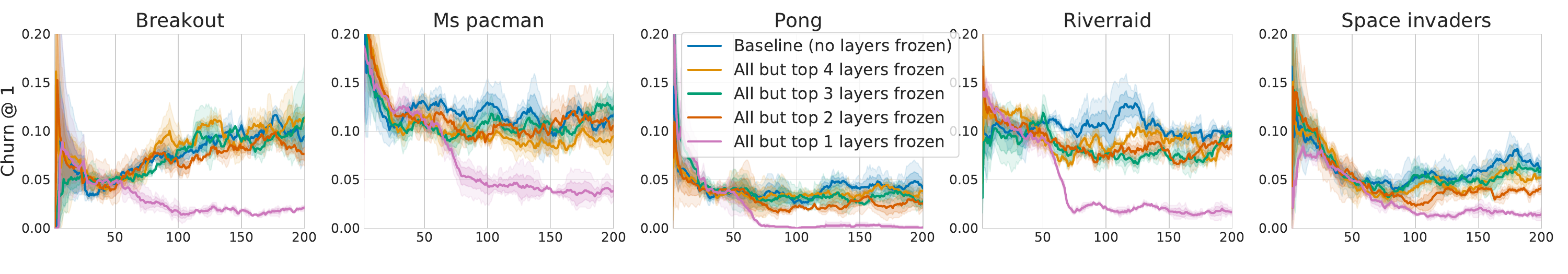}
    \includegraphics[width=\textwidth]{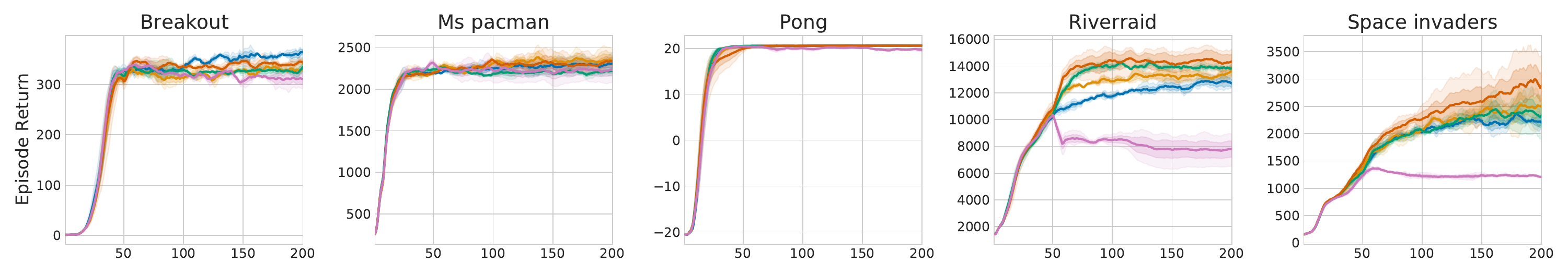}
    \caption{Ablation experiment that relates the depth of the neural network being trained to the amount of policy change.
After $50$M frames of regular training, all but a few top layers of DoubleDQN’s neural network are frozen, and the remainder of training can only change weights in the last $1-4$ layers. 
We find a correlation between churn and trainable capacity, but the most significant step-change occurs between one or more trainable layers, i.e., between linear FA (on top of frozen features) and deep learning.}
    \label{fig:num-frozen-layers}
\end{figure}

\begin{figure}[ptb]
    \centering
    \includegraphics[width=\textwidth]{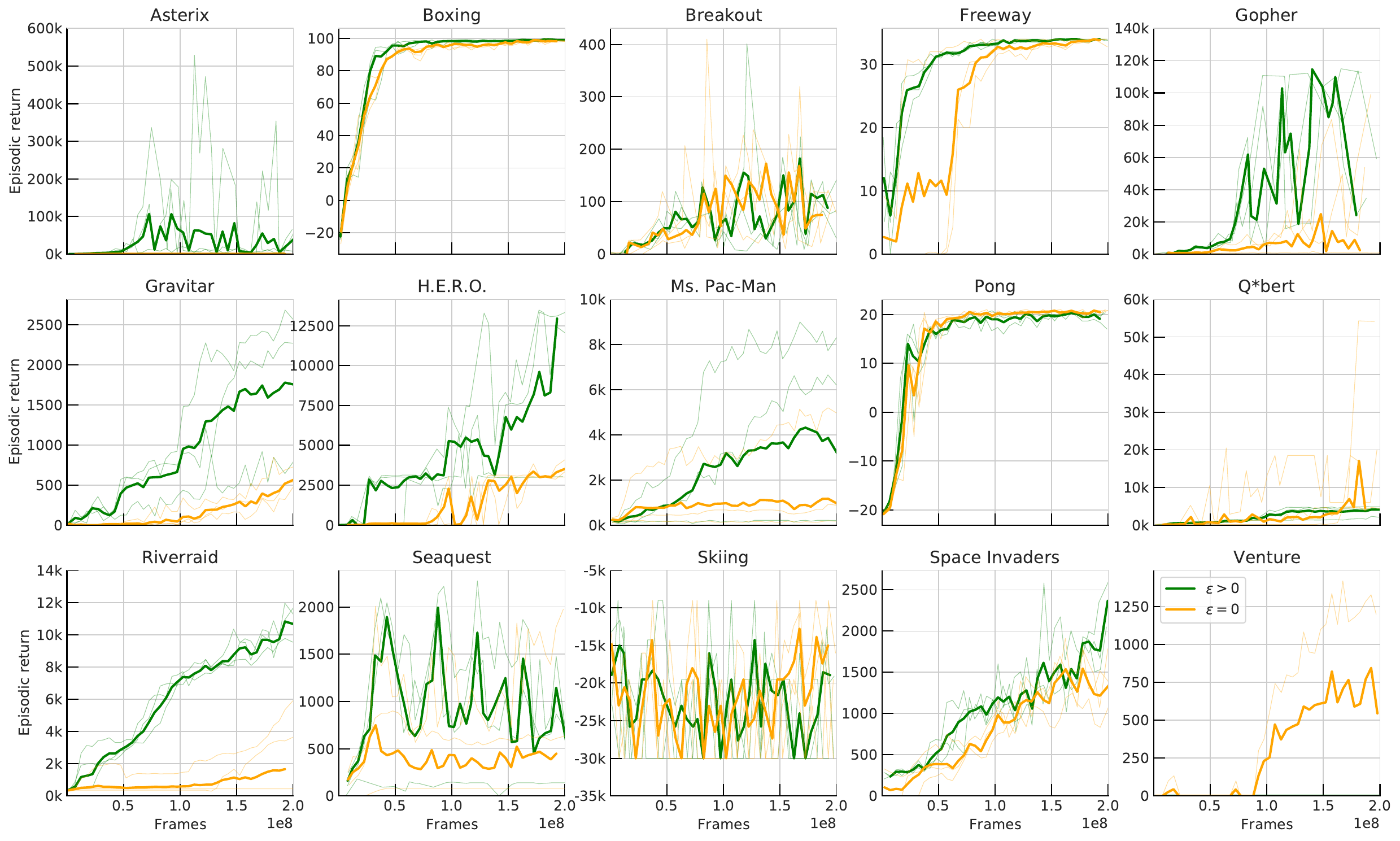}
    \caption{R2D2 performance curves. The setting is the same as in Figure~\ref{fig:zero-eps}, namely $200$M frames, $15$ games, $3$ seeds each (thin lines), but the agent architecture is very different (see Table~\ref{tab:dqn-r2d2}). In comparison, the R2D2 agent is less robust to $\epsilon=0$; despite high policy change, exploration appears to suffer in half of the games.
    We assume this difference is mainly due to two aspects: first, DoubleDQN has a high amount of random exploration in early learning (it takes $4$M frames until $\epsilon$ has decayed to $0$). Second, DoubleDQN traverses many more distinct policy networks over the course of its lifetime ($\approx 10^7$), compared to R2D2 ($\approx 10^4$), due to the latter's much larger batch size, greater parallelism, and smaller replay ratio. Note also that the maximal ``policy age'' (in gradient updates) and as a consequence policy diversity representend in the replay buffer data is very different in R2D2 and DQN. Because of the data generation parallelism (and the near-deterministic dynamics of the Atari environment), diversity of replay data in R2D2 may be driven more by $\epsilon$-exploration than in DQN. The case $\epsilon=0$ may therefore result in a very narrow data distribution and potentially collapse of the neural network representation in R2D2.}
    \label{fig:r2d2-performance}
\end{figure}

\begin{figure}[tbp]
    \centering
    \includegraphics[width=\textwidth]{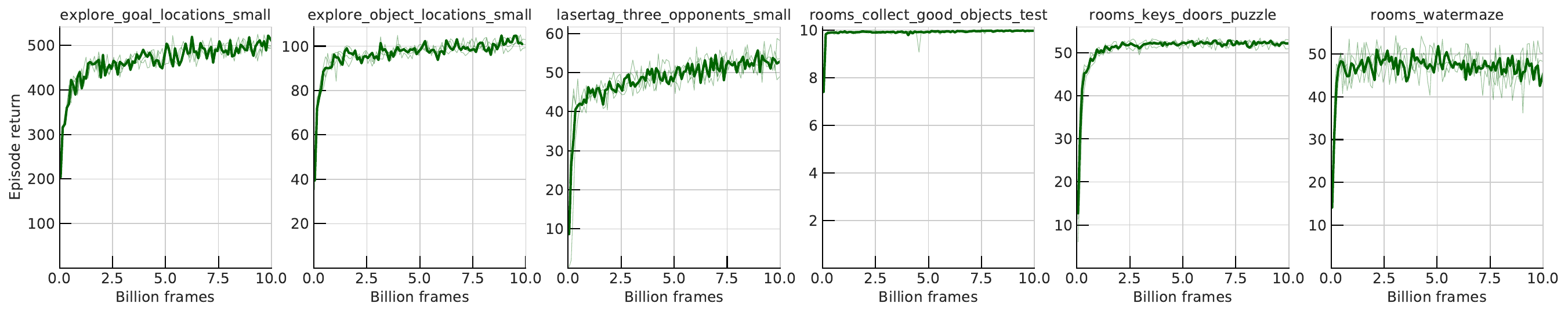}
    \includegraphics[width=\textwidth]{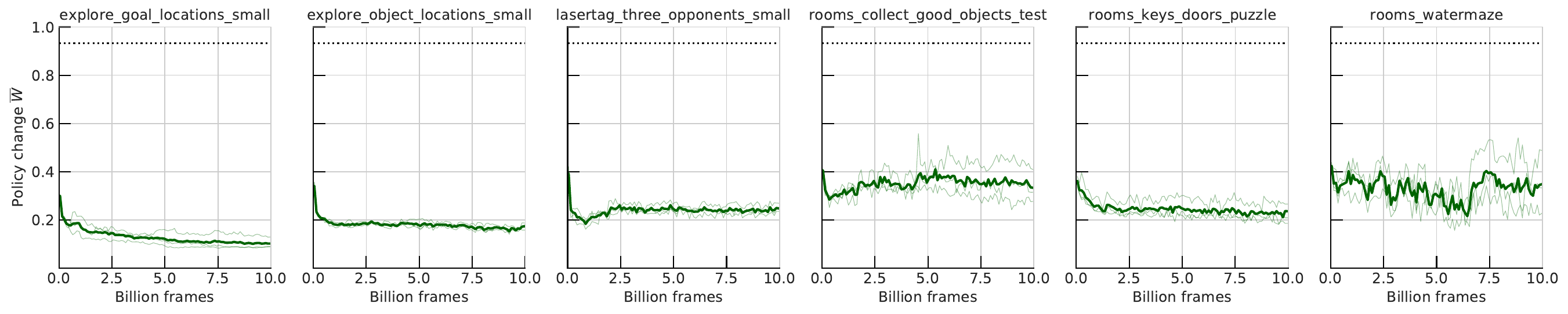}
    \caption{Experiments using R2D2~\cite{r2d2} on $6$ levels of the DM-Lab~\cite{beattie2016deepmind} benchmark suite of 3D environments ($3$ seeds of $10$B frames each), with $|\mathcal{A}|=15$. The average observed policy change ($\approx20\%$, see bottom plots) is overall in line with the Atari results, but somewhat higher, possibly because of the different action space, where most actions can be easily undone at the next step.
}
    \label{fig:dmlab}
\end{figure}

\begin{figure}[tbp]
    \centering
    \includegraphics[width=\textwidth]{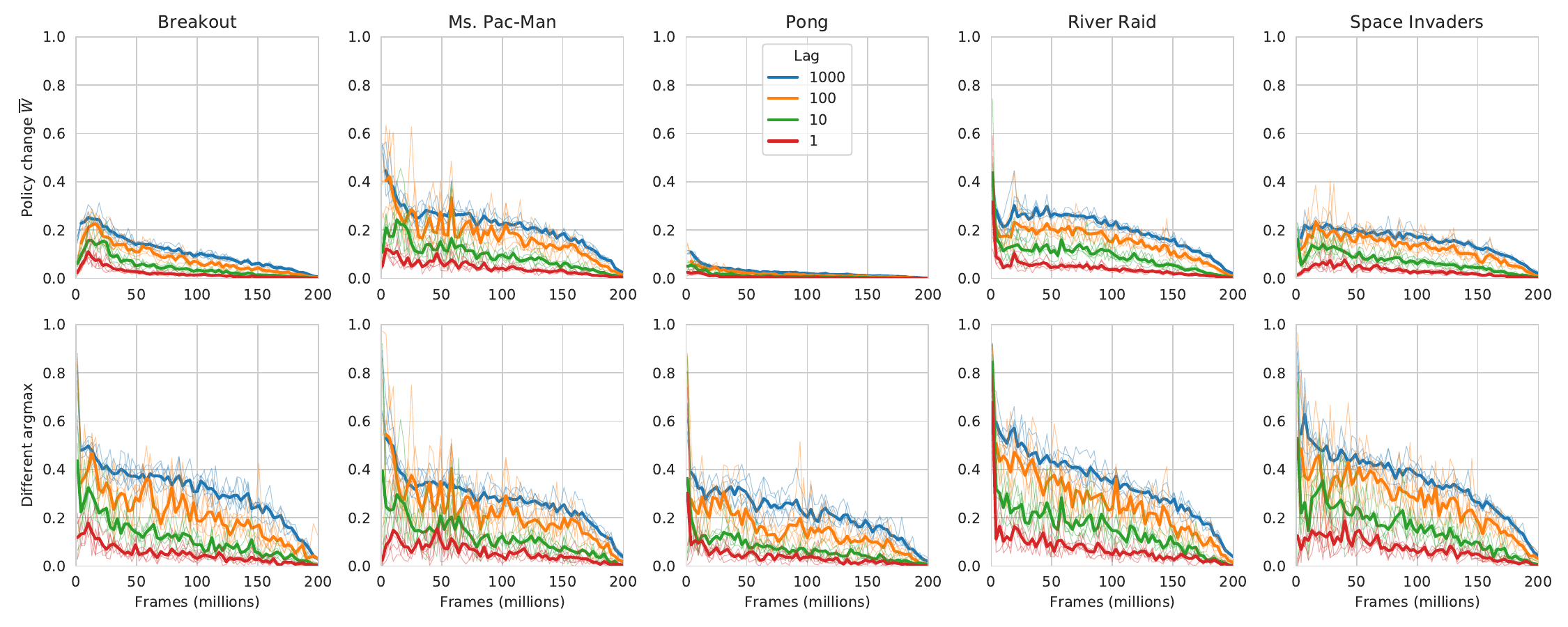}
    \caption{
    Preliminary experiments with an {\bf actor-critic} agent on a subset of Atari games ($5$ seeds each), with minimal action spaces per game (as in the DDQN setup, see Figure~\ref{fig:typical-churn}). The agent is an implementation of IMPALA~\cite{impala} in the Sebulba framework~\cite{hessel2021podracer}. The policy change is comparable to other agents on Atari, showing policy churn is present in actor-critic agents, not just value-based agents. Note policy change reduces to zero as training progresses because the learning rate is linearly annealed to zero.
    {\bf Top row}: Total variation policy change $\overline{W}$, as defined in Eqs.~\ref{eq:pi-change} and~\ref{eq:w-bar}; as these are soft policies, the change is expected to be smaller than it would be for switches between greedy policies. {\bf Bottom row}: Shows the average $\arg\max$ switches on the same experiment. Different colours show different intervals $k$ across which $\W(\pi_t,\pi_{t+k})$ is measured (as in Figure~\ref{fig:typical-churn}).
}
    \label{fig:actor-critic}
\end{figure}

\end{document}